\documentclass{article}

\usepackage[preprint]{neurips_2026}

\usepackage[utf8]{inputenc} 
\usepackage[T1]{fontenc}    
\usepackage{hyperref}       
\usepackage{url}            
\usepackage{booktabs}       
\usepackage{amsfonts}       
\usepackage{nicefrac}       
\usepackage{microtype}      
\usepackage{xcolor}         

\usepackage{graphicx}
\usepackage{amsmath}
\usepackage{multirow}

\title{Gender Artifacts from Art History to Text-to-Image Generation}

\author{%
  Piera Riccio \\
  University of Amsterdam\\
  Amsterdam, Netherlands \\
  \texttt{p.riccio@uva.nl} \\
  \And
  Miriam Doh \\
  Université Libre de Bruxelles \\
  Brussels, Belgium \\
  \And
  Benedikt Höltgen \\
  Hasso Plattner Institut \\
  University of Potsdam \\
  Potsdam, Germany
  \And
  Noa Garcia \\
  The University of Osaka \\
  Osaka, Japan
  \And
  Nanne van Noord \\
  University of Amsterdam \\
  Amsterdam, Netherlands
}

\begin{document}

\maketitle

\begin{abstract}
Artistic styles are rooted in specific socio-historical contexts that encode social hierarchies, including distinct constructions of gender. Yet in AI research, style has long been treated as a surface-level visual property: a filter of color, brushstroke, and texture applied to otherwise content-neutral scenes. We introduce the first dataset to investigate the interplay between gender representation and style in both historical and generated images. \textsc{StyleGender} comprises 74k
images spanning 19 artistic styles, comprising art historical images with style and gender annotations, T2I-generated images under controlled style and gender prompts, and a semantically aligned set enabling direct art history-to-generation comparison. 
By proposing two Set Gender Artifact (SGA) metrics (\textsc{PixelSGA} and \textsc{MaskSGA}), capturing gender signals at the pixel level and in compositional structure, we show that (1) gender representation shapes visual features across artistic styles, (2) style keywords carry these patterns into T2I generation, and (3) generative models tend to amplify gender artifacts beyond what is observed in historical sources.
\end{abstract}

\section{Introduction}
\label{sec:intro}

\begin{figure}[h]
  \centering
  \includegraphics[width=\textwidth]{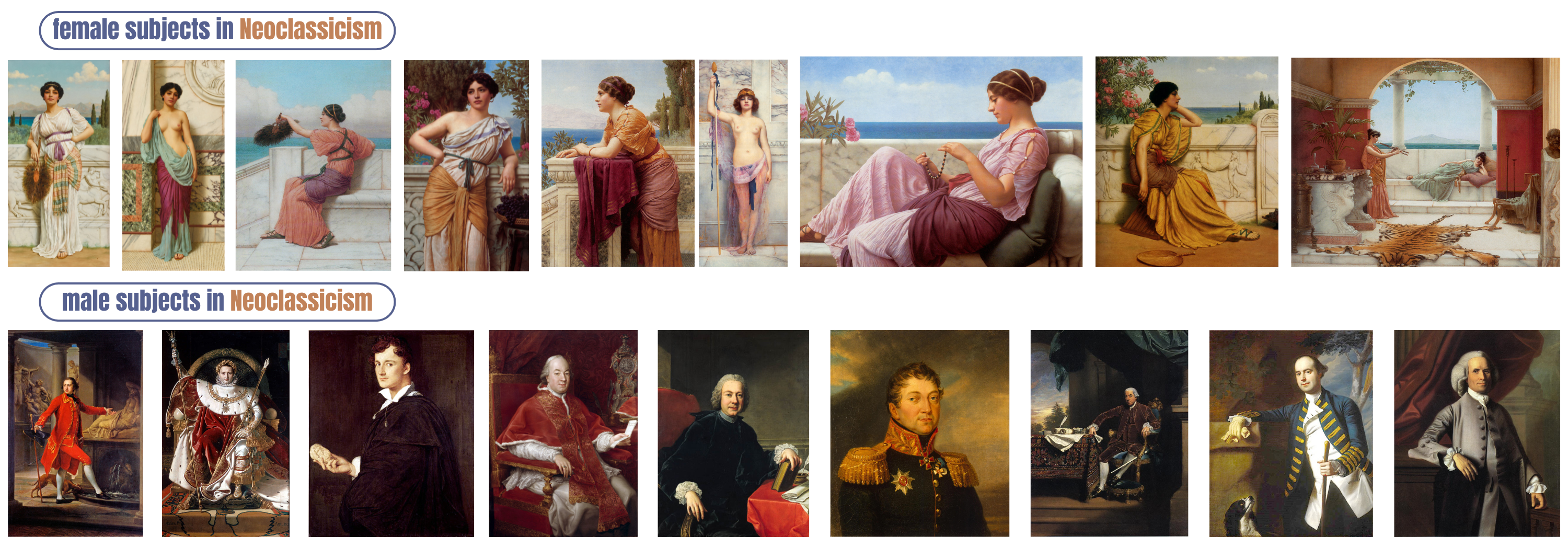}
  \caption{Sample Neoclassicism artworks, illustrating systematic visual differences in setting, color palette, clothing, and pose between female (top) and male (bottom) subjects.} 
  \label{fig:fig1_art_history}
\end{figure}

Artistic styles are not visual templates that emerge in a vacuum. They are rooted in specific socio-historical contexts \cite{crow2004rise,belting2022anthropology,schapiro1994theory}, shaped by cultural norms, institutional structures, and deeply embedded social hierarchies \cite{thiel2016gentleman}. Art historical periods encode distinct constructions of social roles, such as gender \cite{l2016representing,rose2015receptions}. The examples of Neoclassical paintings in Figure \ref{fig:fig1_art_history} demonstrate how salient differences in scene colors (\textit{e.g.}, bright vs. dark), settings (\textit{e.g.}, indoor vs. outdoor), and social roles are apparent depending on the gender of the depicted subjects. 

However, in AI research, the notion of artistic style has been associated primarily with color palettes, brushstrokes, and texture \cite{Kotovenko_2019_ICCV}. This understanding became particularly prominent with the rise of neural style transfer methods \cite{jing2019neural} and early image-to-image translation models such as CycleGAN \cite{zhu2017unpaired}. 
This operational definition has not evolved even with the emergence of text-to-image (T2I) models such as DALL-E \cite{ramesh2021zero}, Stable Diffusion \cite{rombach2022high}, Midjourney \cite{midjourney2022} and FLUX \cite{flux2024}, despite these models generating both aesthetic and semantic content when prompted with explicit references to artistic styles (\textit{e.g.}, “in the style of Impressionism”) \cite{civiverse,diffusiondb}. If T2I models learn the concept of style from art historical images, then prompting a specific historical style may implicitly capture also the social structures embedded in the geo-historical period that produced it. Despite this, style is still often described as a sort of visual filter applied to otherwise content-neutral scenes \cite{wasielewski2024reification}. 

Following the understanding that artistic styles encode differentiated representations of gender, we introduce \textsc{StyleGender}, a dataset designed to study the intersection of artistic style and gender. It comprises around 74k images, in particular: (1) $\sim$18k art historical images spanning 19 historical styles and 2 gender labels; (2) 19k images generated by two T2I models (Stable Diffusion 3.5 Medium and Flux.1-dev) using simple prompts varying only the gender and style keywords; and (3) a semantically aligned set in which each of the 18k art historical images is re-generated using the two T2I models. We conduct an empirical analysis on this dataset that builds on the definition of a ``gender artifact'' proposed by Meister et al. \cite{meister2023gender} — “\textit{a visual cue that is correlated with gender, [...] learnable by a modern image classifier and [that] have an interpretable human corollary}”. In particular, we introduce two novel metrics to quantify gender artifacts in sets of images, which we define as Set Gender Artifacts (SGA): PixelSGA and MaskSGA. 
Our findings reveal that artistic styles in art history exhibit clear gender artifacts across both pixel features and compositional patterns, with Neoclassicism and Art Nouveau scoring highest. Moreover, we find that the outputs of two analyzed T2I models also encode gender artifacts, and that varying stylistic keywords systematically modulates these patterns. While the two T2I models differ in how they distribute these gender cues across styles, both tend to amplify pre-existing artifacts relative to historical works. 

Overall, our evaluation demonstrates that artistic styles in AI research should be reframed as socio-cultural constructs rather than purely aesthetic transformations.

\section{Related Work}

Torralba and Efros \cite{torralba2011unbiased} were among the first to systematically characterize the existence of biases in visual datasets, showing that models trained on a given dataset generalize poorly across others. Fabbrizzi et al. \cite{fabbrizzi2022survey} provide a taxonomy of how bias enters visual datasets through selection, framing, and labeling processes, and how this can also result in social biases along protected attributes \cite{yang2020towards,birhane2021large,garcia2023uncurated}. Among these, gender bias is one of the most extensively studied, appearing across a wide range of visual recognition tasks \cite{buolamwini2018gender,barlas2021see,hirota2022quantifying,yang2022enhancing}. Meister et al. \cite{meister2023gender} showed that gender-correlated cues (\textit{i.e.}, gender artifacts) are pervasive in large-scale visual datasets and largely infeasible to remove.

Gender biases have been shown to persist in generative models and, in particular, T2I systems reflect and amplify societal biases present in their training data \cite{chinchure2024tibet,luccioni2023stable,bianchi2023easily,wilson2025bias}. This phenomenon has been widely documented across multiple dimensions \cite{wan2024survey,elsharif2025cultural}. One line of work examines bias in default generation: models often depict a particular gender even for neutral prompts such as “person” or “a face” \cite{zhang2023iti,garcia2023uncurated,wu2024stable}. Another line of work considers stereotypical association bias, where models project culturally encoded gender roles onto prompt content. Occupational bias is a prominent example, with models over- or under-representing genders for profession-related prompts \cite{wang2024new,mannering2023analysing,luccioni2023stable}, quantified via output gender distributions \cite{zhou2024biasgenerativeai,sun2024smiling} or embedding proximity in the model's latent space \cite{wang2023t2iat,mandal2023multimodal}. Similar patterns appear for personal traits and interests, linking men to “science” or career contexts and women to “art” or family domains \cite{wang2023t2iat}, and representing social attributes like friendliness or trustworthiness with normative aesthetics (young, smiling, groomed), often amplified for women \cite{fraser2023friendlyfacetexttoimagesystems, luo2024bigbench, doh2026aestheticsstructuralharmalgorithmic}. Gender-coded objects also co-occur with prompts even when unprompted: male prompts elicit ties or baseball bats, female prompts handbags or domestic items \cite{mannering2023analysing, bansal2022well}. Bias further manifests in model behavior beyond prompt semantics, including disparities in image quality, harmful content generation, and sexualization of women and non-cisgender subjects \cite{qu2023unsafe,ungless2023stereotypes,hao2024harm,doh2026aestheticsstructuralharmalgorithmic}. At the data level, audits of large-scale corpora such as LAION-400M reveal systematic reinforcement of misogynistic associations, with sexualized depictions of women appearing even for semantically neutral queries \cite{birhane2021multimodaldatasetsmisogynypornography,birhane2023into}. Yet the role of artistic style as an independent carrier of gender-coded visual patterns remains largely unexplored, motivating our investigation.
 
Rather than being understood as culturally and historically situated, artistic styles in AI research are indeed often treated as separable from content, with research exploring mechanisms to disentangle them \cite{Kotovenko_2019_ICCV}, though recent work shows that style and content remain at least partially entangled \cite{wu2024goya}. Early neural style transfer approaches followed a similar perspective, modeling style primarily through low-level visual elements—such as color distributions, texture, and brushstroke patterns—while preserving high-level content \cite{jing2019neural}, effectively treating style as a transformation of appearance. In addition, the literature presents efforts both to generate stylistic novelty \cite{elgammal2017can} and to quantitatively evaluate faithfulness in stylistic replication \cite{wright2022artfid}. In T2I generation, prompts provide a natural mechanism for controlling style, and specifying stylistic keywords has been shown to influence not only low-level visual features but also semantic and compositional aspects of the outputs \cite{chen2023controlstyle, app151910623, schaerf2025training}. In particular, specifying a style in a T2I model reproduces widely represented, iconic exemplars, suggesting that generative models encode artistic style in a decontextualized, statistical manner \cite{wasielewski2024reification}.

In such a context, our research emphasizes that artistic styles should not be treated as neutral visual elements, but rather as carriers of social and cultural meaning.

\section{The \textsc{StyleGender} Dataset}
\label{sec:dataset}

The \textsc{StyleGender} dataset\footnote{Dataset at: \url{https://huggingface.co/datasets/StyleGender/StyleGender-Dataset}.} is designed to study interactions between gender representation and artistic style, and comprises images annotated with gender and style labels. It is constructed from three data sources: (1) art historical paintings from the WikiArt online collection, (2) images generated by two T2I models using controlled prompts varying gender and style, and (3) images generated to be semantically aligned with the art historical samples. In this dataset, gender is represented as a binary label (male, female), based on available annotations in art historical images and explicitly specified prompts in generated images. These labels reflect perceived or expressed gender in prompt generation, and are used solely as an analytical tool to investigate stylistic patterns, without making claims about personal identity. We next describe the three sources of \textsc{StyleGender}.

\begin{figure}[tb]
  \centering
  \includegraphics[width=0.9\textwidth]{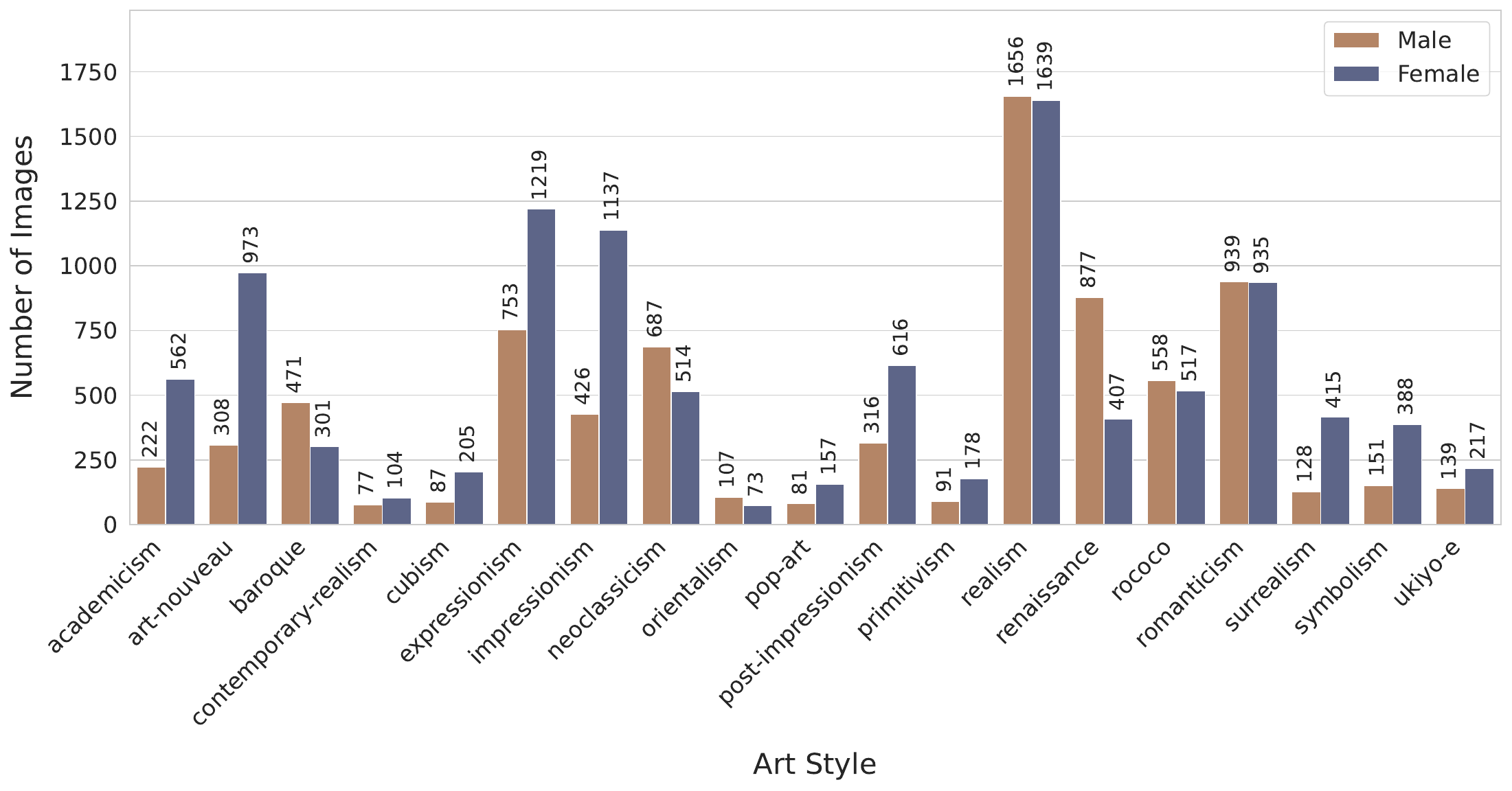}
  \caption{Distribution of male and female artworks across the 19 artistic styles in our curated sample of art historical images, comprising 8,074 male and 10,557 female images.
  }
  \label{fig:wiki_distributions}
\end{figure}

\paragraph{Art History}
We rely on the WikiArt online collection to obtain a large-scale sample of historical artworks in different styles. We assign gender labels based on male- and female-related tags using the \textit{advanced search} functionality of WikiArt. We then manually curate the dataset, excluding images without human subjects and images that clearly include at least one subject of the opposite gender. Details on this process are provided in Appendix \ref{sec:apx:wiki}.
The resulting dataset comprises 18,631 artworks spanning 19 styles, with 8,074 male and 10,557 female subjects. Sample sizes vary across styles (Figure~\ref{fig:wiki_distributions}), with some styles (\textit{e.g.}, Realism, Romanticism) more populated than others (\textit{e.g.}, Orientalism, Contemporary Realism). 

\paragraph{T2I Generation}

\begin{figure}[tb]
  \centering
  \includegraphics[width=\textwidth]{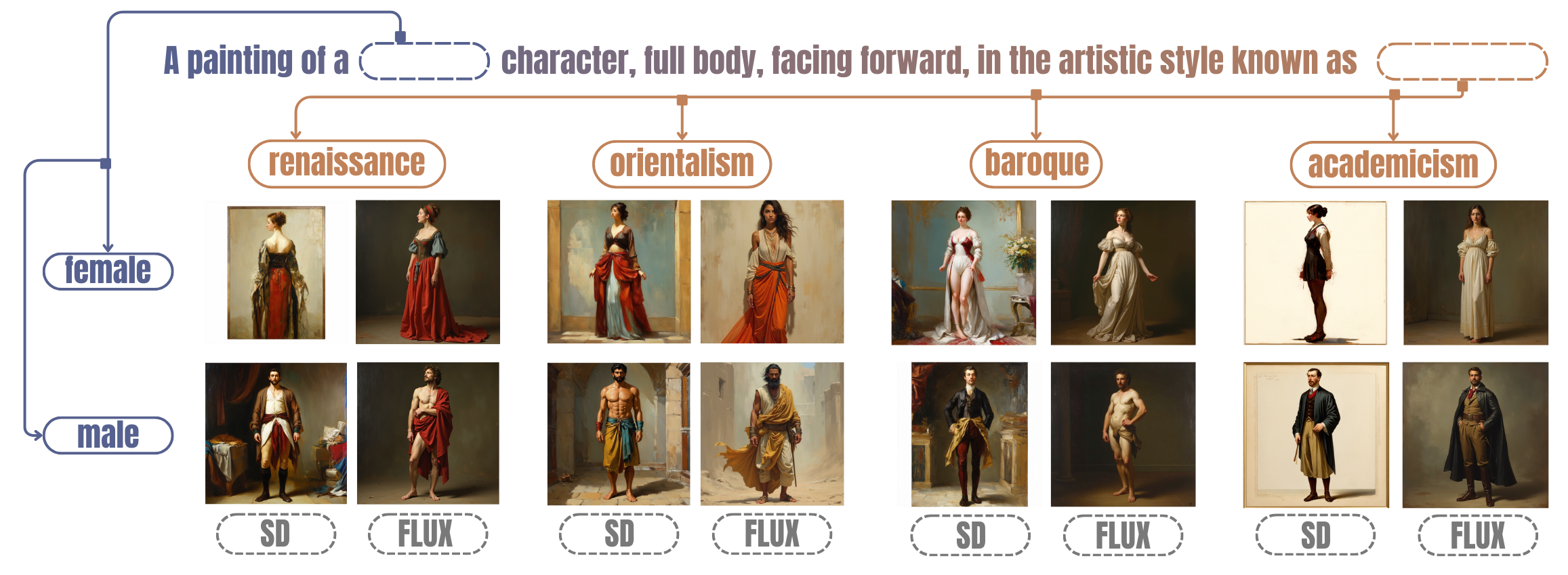}
  \caption{Sample images generated with Stable Diffusion 3.5 Medium (SD) and Flux.1-dev (Flux) using the prompt template “\texttt{A painting of a \textcolor{blue}{<gender>} character, \textcolor{teal}{full body}, \textcolor{magenta}{facing forward}, in the artistic style known as \textcolor{orange}{<style>}}" across four artistic styles.} 
  \label{fig:rq2_method}
\end{figure}

We use two open-source T2I models: Stable Diffusion 3.5 Medium (\emph{SD}) \cite{esser2024scaling,stabilityai_sd35_medium_2024} and Flux.1-dev (\emph{Flux}) \cite{flux1_dev}. For each of the 19 artistic styles under consideration, we generate 500 images per model (250 male, 250 female). To ensure comparability across styles, we use controlled prompts specifying (i) gender, (ii) viewpoint, (iii) pose, and (iv) artistic style, following the template: \texttt{A painting of a \textcolor{blue}{<gender>}, \textcolor{teal}{<viewpoint>}, \textcolor{magenta}{<pose>}, in the artistic style known as \textcolor{orange}{<style>}}, \textit{e.g.}, \textit{A painting of a \textcolor{blue}{female} character, \textcolor{teal}{full body}, \textcolor{magenta}{facing forward}, in the artistic style known as \textcolor{orange}{Baroque}}. For these prompts, gender terms are drawn from a controlled vocabulary: females as ``woman'', ``female character'', or ``human female''; males as ``man'', ``male character'', or ``human male''. Pose and viewpoint variations are systematically combined to introduce controlled diversity. For \emph{Pop Art}, we slightly modify the prompt to \texttt{A \textcolor{blue}{<gender>}, \textcolor{teal}{<viewpoint>}, \textcolor{magenta}{<pose>} in \textcolor{orange}{<style>} style}, because referring to “painting” yields stylistically weaker outputs for this category, which frequently appears as graphic or digital art. The full prompt set is provided in Appendix \ref{sec:apx:prompts} and exemplary generations are shown in Figure \ref{fig:rq2_method}. The figure illustrates that modifying only the stylistic keyword leads not only to visual changes in texture and color, but also to differences in higher-level semantic elements of the generated images, such as scene setting or clothing.

\paragraph{From Art History to T2I Generation}

\begin{figure}[tb]
    \centering
    \includegraphics[width=\linewidth]{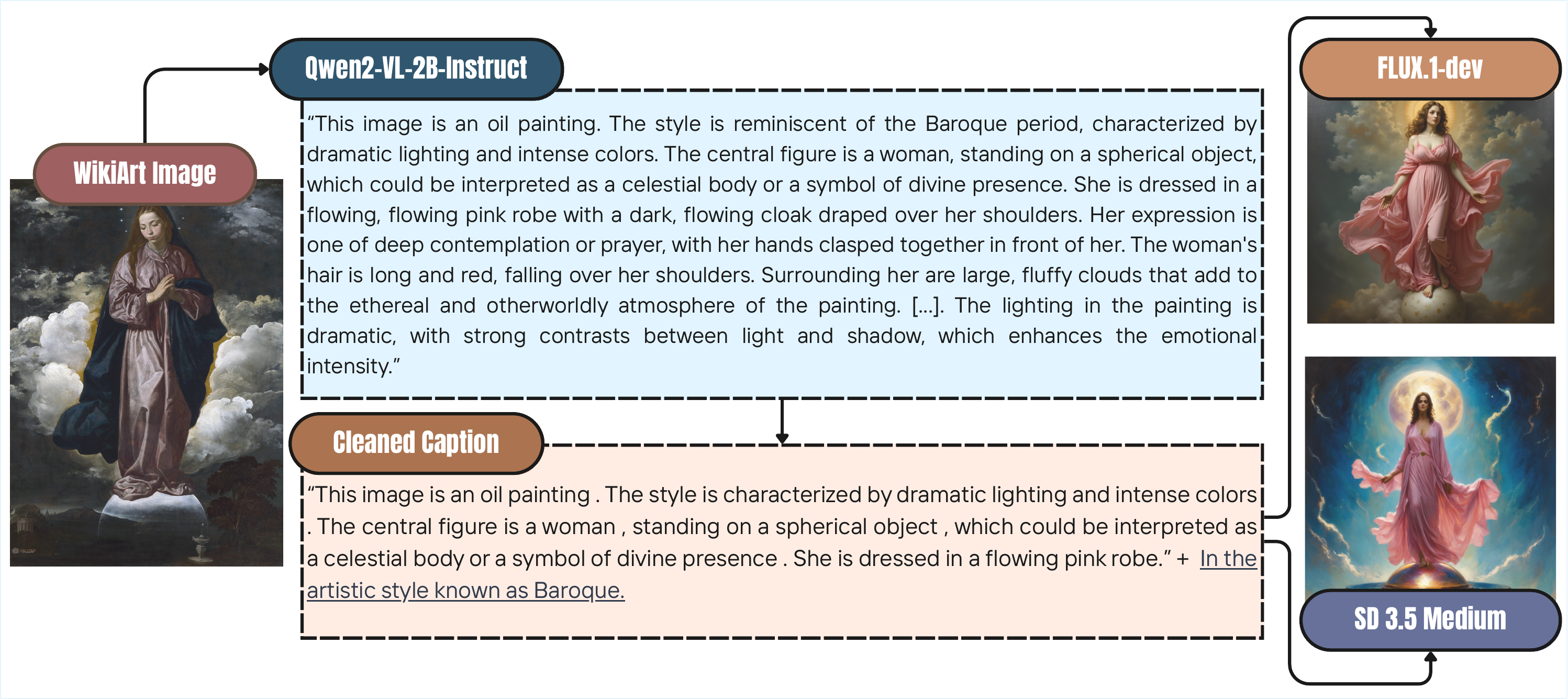}
    \caption{The ``Immaculate Conception'' by Diego Velázquez is first captioned by Qwen2-VL-2B-Instruct. The caption is cleaned and truncated to 60 tokens; the artistic style is re-appended as a controlled suffix. The resulting prompt is used to generate semantically aligned images.}
    \label{fig:qwen}
\end{figure}

We generate a second set of images using prompts that, in addition to the stylistic conditioning, are also semantically aligned with the art historical images. 
First, textual captions for the historical images are generated using Qwen2-VL-2B-Instruct \cite{Qwen2VL}, prompted to describe visible content only (medium, subjects, clothing, expressions, poses, scene elements, colors, lighting) and avoid references to style, art movements, or artists. The captions are cleaned via a regex filter removing style/period mentions, truncated to 60 CLIP tokens to leave space for adding the right stylistic keyword explicitly, and corrected to enforce gender consistency with the source artwork, replacing neutral terms (\textit{e.g.}, “figure”) with explicit gender labels. Rare truncation errors were manually corrected, as detailed in Appendix \ref{sec:apx:qwen}. These captions are then used to generate images with \textit{SD} and \textit{Flux}. We assess the effectiveness of this approach qualitatively, with examples provided in  Figure~\ref{fig:qwen} and Appendix \ref{sec:apx:qwen}. The generated captions and the curated prompts are also part of the \textsc{StyleGender} dataset.

\section{The Set Gender Artifacts (SGA) Metrics}
\label{sec:metric}

The \textsc{StyleGender} dataset enables analyses of how gender representations interact with artistic styles. As a foundational step, we evaluate correlations between gender and style in terms of pixel-level and compositional features, providing the basis for subsequent higher-level analyses. For doing so, we rely on previous work \cite{meister2023gender}, showing the existence of gender artifacts in image datasets. 
Building on their approach, we introduce two Set Gender Artifacts (SGA) metrics, namely PixelSGA and MaskSGA,\footnote{Code at: \url{https://anonymous.4open.science/r/GenderArtifacts_ArtHistory_T2I-6C80/README.md}.} which allow to measure and compare the extent to which gender-discriminative signals are embedded within sets of images. In both cases, we evaluate how much gender information remains after systematically removing relevant visual information, based on the performance of a gender classifier. Gender classification is employed in this work solely to compute the \textsc{SGA} metrics and is not used for downstream purposes.  
We formally define these measures as follows.


Consider a dataset $\mathcal{D}$ consisting of labeled images $(x_i, y_i) \in \mathcal{X} \times \mathcal{Y}$ where each image $x_i$ depicts one or more subjects of the same gender, denoted by $y_i \in \mathcal{Y}$.
Let $N_y$ denote the number of samples with gender $y \in \mathcal{Y}$. We then denote the true positive rate of a classifier $h : \mathcal{X} \to \mathcal{Y}$ for class $y$ on $\mathcal{D}$ by
\begin{equation}
\mathrm{TPR}^{y}(\mathcal{D}) := \frac{1}{N_y} \sum_{i:y_i=y} \mathbf{1}[h(x_i) = y].
\end{equation}
The \emph{balanced accuracy} on $\mathcal{D}$ is the mean of the per-class true positive rates:
\begin{equation}
\mathrm{BA}(\mathcal{D}):
= \frac{1}{|\mathcal{Y}|} \sum_{y \in \mathcal{Y}} \mathrm{TPR}^{y}(\mathcal{D}).
\end{equation}

\paragraph{\textsc{PixelSGA}}

\begin{figure}[tb]
  \centering
  \includegraphics[width=\textwidth]{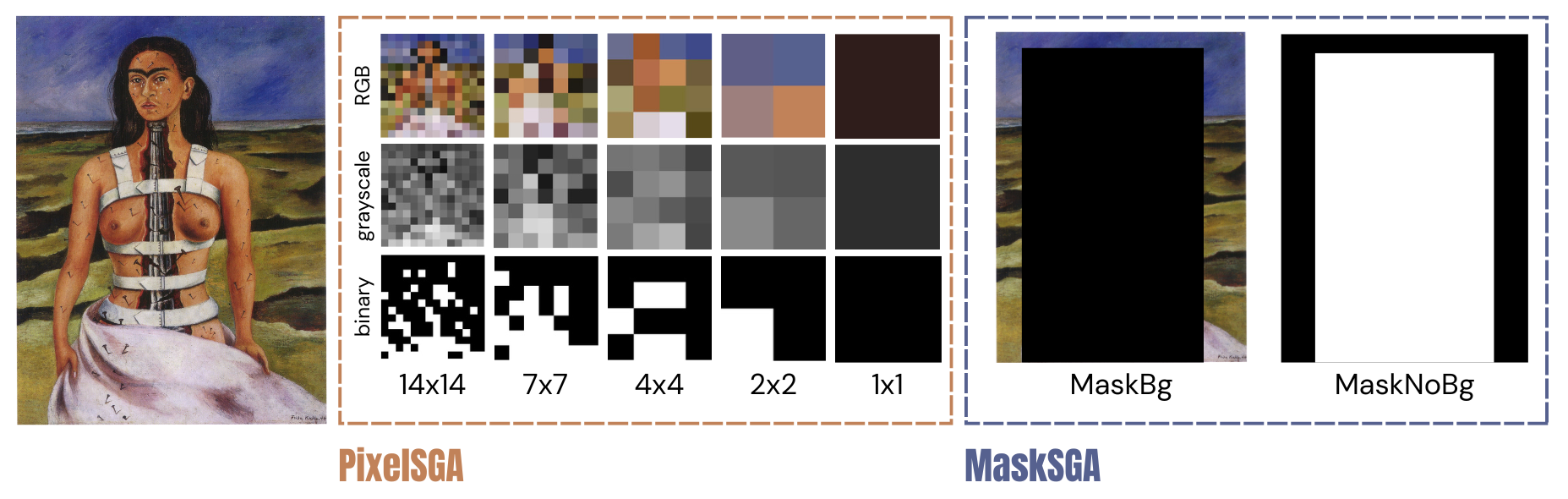}
  \caption{On the left-hand side, "The broken column" by Frida Khalo. In the center, exemplary manipulations on which \textsc{PixelSGA} can be computed. On the right-hand side, exemplary manipulations on which \textsc{MaskSGA} can be computed.
  } \label{fig:sga_manipulations}
\end{figure}

To quantify pixel-level gender cues, we focus on manipulated images generated via spatial downsampling across chromatic encodings (examples in Figure \ref{fig:sga_manipulations}). 
Given an image $x \in \mathcal{X}$, we define downsampled versions $x^{(r)}$ with resolutions $r \in \mathcal{R}$.
For low resolutions, the representation retains only coarse structural or global color information. For $1 \times 1$, the image collapses to a single pixel, encoding the global average color.
Low-level gender artifacts may differ in which chromatic encoding, such as RGB or binary, makes them most salient.
Therefore, one may consider multiple versions $x^{(r, \tau)}$ of each downsampled image $x^{(r)}$, with $\tau \in \mathcal{T}$ denoting the chromatic encoding.
This procedure yields $|\mathcal{R}| \times |\mathcal{T}|$ manipulated variants $x^{(r, \tau)}$ per image, and equally many resulting datasets $\mathcal{D}^{(r, \tau)}$. At each resolution $r$, we take the maximum BA over chromatic encodings $\tau$ to capture encoding-invariant gender separability. Averaging over resolutions yields:

\begin{equation}\label{eq:pixelSGA}
\textsc{PixelSGA}: = \frac{1}{|\mathcal{R}|} \sum_{r \in \mathcal{R}} \max_{\tau \in \mathcal{T}}\ \mathrm{BA}(\mathcal{D}^{(r, \tau)}).
\end{equation}
\textsc{PixelSGA} takes values in $[\frac{1}{|\mathcal{Y}|}, 1]$ unless the classifier performs worse than random; higher values indicate stronger gender-discriminative signals across low-level visual encodings.

\paragraph{\textsc{MaskSGA}}

To measure gender-discriminative information in compositional structure or contextual background, for each image $x$, we apply masks that cover all human subjects.
We then construct two manipulated versions $x^{(\text{Mask})}$ and $x^{(\text{MaskNoBg})}$ by occluding either only the masked area (Mask) or both the mask and the background (MaskNoBg) using different colors. 
Occluding only the mask removes the visual content of the subjects while retaining contextual information, hence classification performance on the resulting $\mathcal{D}^{(\text{Mask})}$ reflects the extent to which gender cues are encoded in background elements.
In contrast, occluding both the mask and the background preserves only information about the size and position of the subjects and the performance on $\mathcal{D}^{(\text{MaskNoBg})}$, therefore, reflects the extent to which gender information is encoded purely in spatial composition (\textit{e.g.}, scale or placement of the subject within the frame). 
Illustrative examples of these manipulations with rectangular masks are provided in Figure~\ref{fig:sga_manipulations}. The average of the balanced accuracies for datasets $\mathcal{D}^{(\text{Mask})}$ and $\mathcal{D}^{(\text{MaskNoBg})}$ defines the \textsc{MaskSGA} score:
\begin{equation}\label{eq:maskSGA}
\textsc{MaskSGA} := \frac{1}{2} \Big(\mathrm{BA}(\mathcal{D}^{(\text{Mask})}) + \mathrm{BA}(\mathcal{D}^{(\text{MaskNoBg})}) \Big),
\end{equation}
\textsc{MaskSGA} also normally takes values in $[\frac{1}{|\mathcal{Y}|}, 1]$, and measures the strength of gender-discriminative cues encoded in spatial layout or contextual background.

\section{Experiments}
\label{sec:experiments}

We compute the SGA metrics on \textsc{StyleGender} by partitioning the dataset based on the data sources described in Section \ref{sec:dataset}. 
Since gender is represented as a binary label, chance-level performance is $0.50$, with values above this threshold indicating the presence of a gender artifact. For \textsc{PixelSGA}, images are downsampled to resolutions $\mathcal{R} = \{14, 7, 4, 2, 1\}$, as prior work shows observers can still recognize gender at $16\times16$ \cite{torralba2009many}. Each downsampled image is transformed into color, grayscale, and binary b/w versions ($\mathcal{T} = \{\text{color}, \text{grayscale}, \text{binary}\}$). For \textsc{MaskSGA}, person bounding boxes are obtained using DINO \cite{zhang2023dino}, masking all boxes with confidence near the highest per image, and validated against manually annotated data (details are provided in Appendix \ref{sec:apx:bb}). Gender classification is performed via leave-one-out k-nearest neighbors (k=5) with majority voting on CLIP embeddings. We use OpenCLIP (via the \texttt{open\_clip\_torch} library) with a ConvNext backbone  
pretrained on the LAION-2B 
dataset. To mitigate the effects of class imbalance, the majority class is randomly subsampled for every image, and all results are based on balanced accuracy. The uneven representation of styles motivates our choice for a kNN gender classifier, avoiding training, which would be unreliable on the styles with few images. 

Because kNN on CLIP embeddings may be unstable under image transformations, we perform a resolution-based sanity check. Since downsampling reduces information, a well-behaved classifier should exhibit monotonically decreasing balanced accuracy with decreasing resolution (as in \textsc{PixelSGA}). We quantify this via Spearman’s rank correlation between resolution and accuracy (per style and experiment), and define an overall \textit{monotonicity score} by averaging the z-transformed $\rho$ across styles. We evaluate resolutions $\mathcal{R}_{val} = \{224, 112, 56, 28, 14, 7, 4, 2, 1\}$ and report this score as a validation in every experiment, where $1$ indicates perfect monotonicity across styles.

\paragraph{Gender Artifacts in Art History}
\label{sec:blind}

\begin{figure}[tb]
  \centering
  \includegraphics[width=\textwidth]{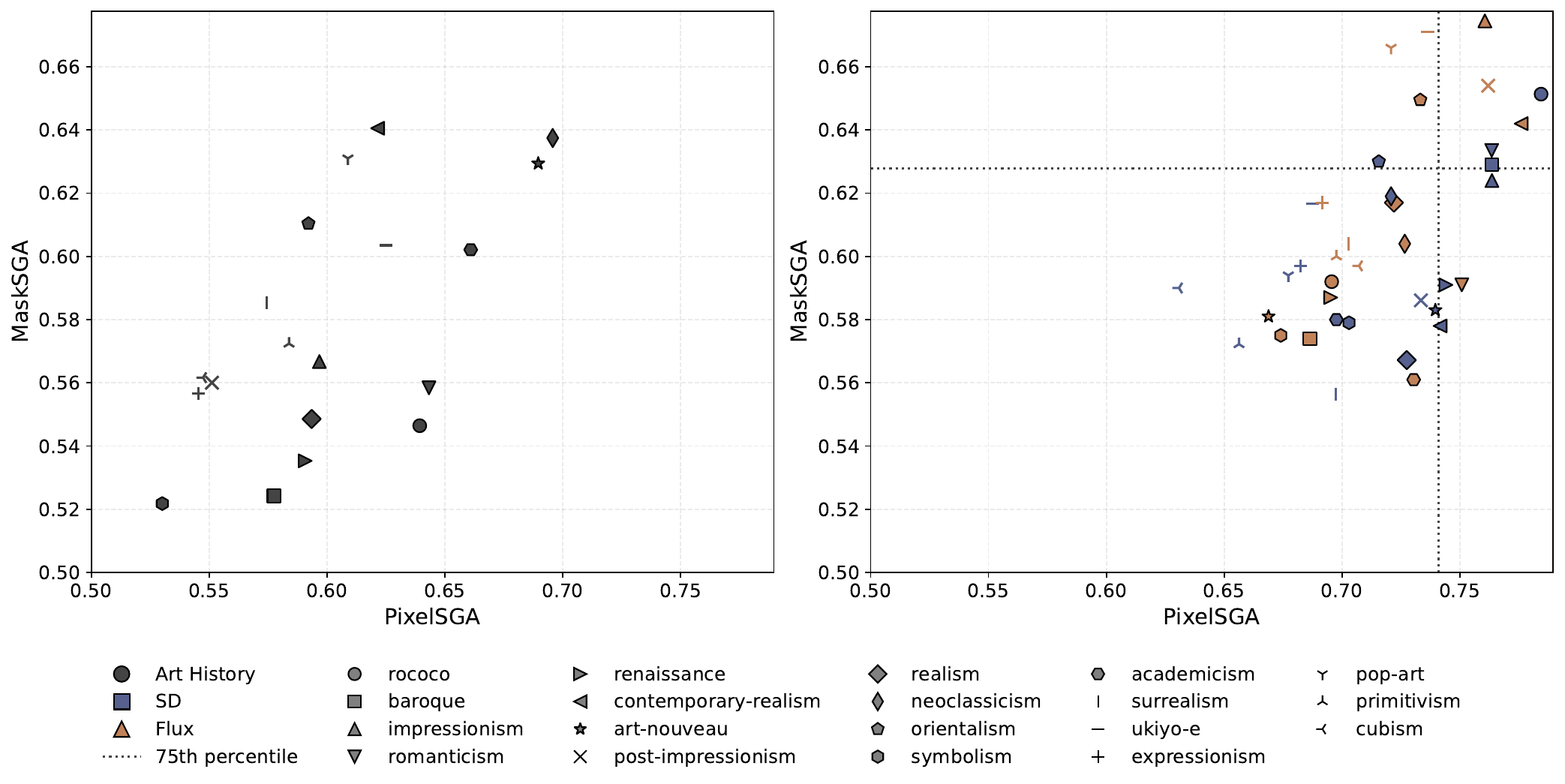}
  \caption{Left: \textsc{PixelSGA} and \textsc{MaskSGA} scores for historic paintings of different art styles. Symbolism (bottom left) shows the least encoding of gender, while Neoclassicism (top right) shows the most. Right: Scores for images generated with \textit{SD} and \textit{Flux} for the same styles. All sets of images show moderate correlation between \textsc{PixelSGA} and \textsc{MaskSGA}.
  }
  \label{fig:rq1}
\end{figure}

We compute \textsc{PixelSGA} and \textsc{MaskSGA} on the art historical images in \textsc{StyleGender} (Figure~\ref{fig:rq1}, left-hand side). All styles show scores above random chance (0.50), indicating the presence of gender artifacts, except Symbolism, whose scores are close to chance ( $\approx$ 0.52). Neoclassicism and Art Nouveau score highest (\textsc{PixelSGA} $\approx$ 0.70/0.69, \textsc{MaskSGA} $\approx$  0.64/0.63), showing that gender information persists even under strong pixelation and masking. Qualitative inspection of the images supports these quantitative findings. Additional results for this experiment are provided in Appendix \ref{sec:apx:wiki_results}. Across styles, \textsc{PixelSGA} and \textsc{MaskSGA} are moderately correlated (Pearson $r=0.63$)  and the \textit{monotonicity score} is equal to $0.998$. 

\paragraph{Gender Artifacts in Text-to-Image Generation}

We further measure how images generated by \textit{SD} and \textit{Flux} under simple prompt structure (Figure \ref{fig:rq2_method}) present gender artifacts. The results in Figure~\ref{fig:rq1} (right-hand side) show that all styles in both models exhibit gender artifacts, and varying the stylistic keyword in prompts produces systematic differences in gender separability. \textsc{PixelSGA} and \textsc{MaskSGA} are moderately correlated (Pearson $r=0.51$ for \textit{SD} and Pearson $r=0.64$ for \textit{Flux}), and the \textit{monotonicity score} equals to $0.997$ for \textit{SD} and $0.998$ for \textit{Flux}.
Styles above the 75th percentile highlight strong gender artifacts: Rococo, Romanticism, and Baroque for \textit{SD}; Impressionism, Post-Impressionism, and Contemporary Realism for \textit{Flux}, with no overlap, suggesting the models encode stylistic gender cues differently. Complete results for this experiment are reported in Appendix \ref{sec:apx:rq2_results}.

\paragraph{From Art History to Text-to-Image Generation}

\begin{figure}[tb]
    \centering
    \includegraphics[width=\linewidth]{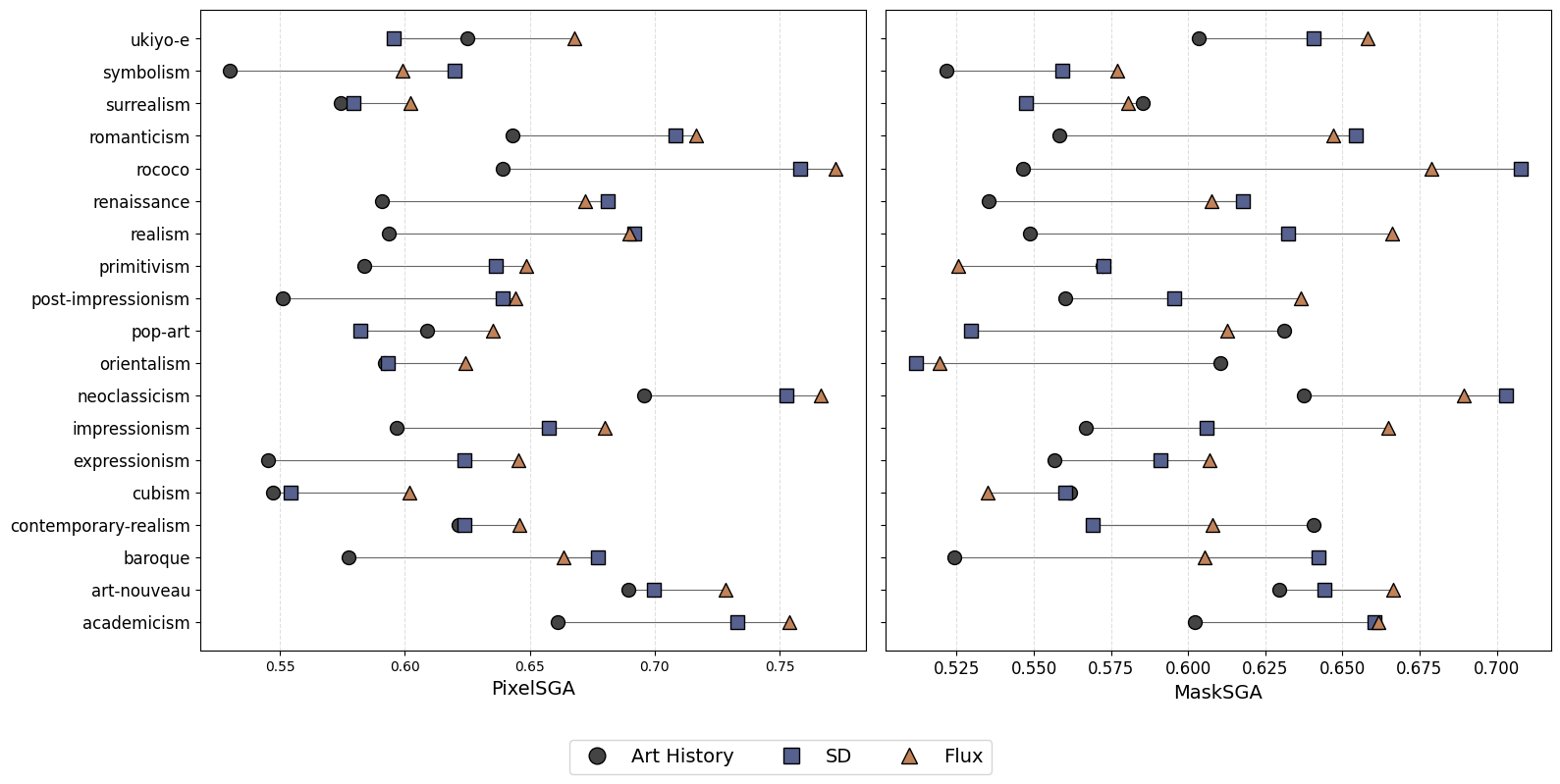}
    \caption{\textsc{PixelSGA} and \textsc{MaskSGA} scores on generated images based on prompts that are semantically aligned with our art history dataset, to enable a cross-domain comparison.} 
    \label{fig:rq3}
\end{figure}

Stereotypical representations in T2I models are concerning not only when present, but also when amplified beyond what is observed in real-world data \cite{dehdashtianoasis}. 
We therefore perform a cross-domain experiment with images generated by \textit{SD} and \textit{Flux} that are semantically aligned with the art historical works (Figure \ref{fig:qwen}). The results in Figure~\ref{fig:rq3} show that \textit{SD} and \textit{Flux} exhibit higher \textsc{PixelSGA} scores than their art historical counterparts in the majority of styles. The effect is more pronounced for \textit{Flux}, where it consistently achieves higher values than the original images.  Rococo stands out as the clearest example of amplification, with both models substantially increasing gender artifact levels. However, the results on the \textsc{MaskSGA} scores present some cases (Surrealism, Pop Art, Orientalism, Cubism and Contemporary Realism) where the art historical values are higher than the generated ones. The MaskSGA scores of the styles with generated images show no correlation with the original styles (Spearman's $|\rho| < 0.2$), but \textit{SD} and \textit{Flux} scores are correlated with a coefficient of $\rho=0.78$, indicating some consistency. Consistency across T2I models is even stronger for PixelSGA, with $\rho=0.92$. On this metric, the images generated by both models also exhibit moderate to strong correlation with the art history, with $\rho=0.61$ (\textit{SD}) and $\rho=0.81$ (\textit{Flux}), respectively. Both models achieve perfect \textit{monotonicity score} ($1.000$). Taken together, these findings suggest that T2I models tend to amplify pre-existing gender artifacts, particularly when it comes to pixel-level features, although the magnitude of this amplification is style-dependent. Complete results reported in Appendix \ref{sec:apx:rq3_results}.

\begin{figure}[tb]
    \centering
    \includegraphics[width=\linewidth]{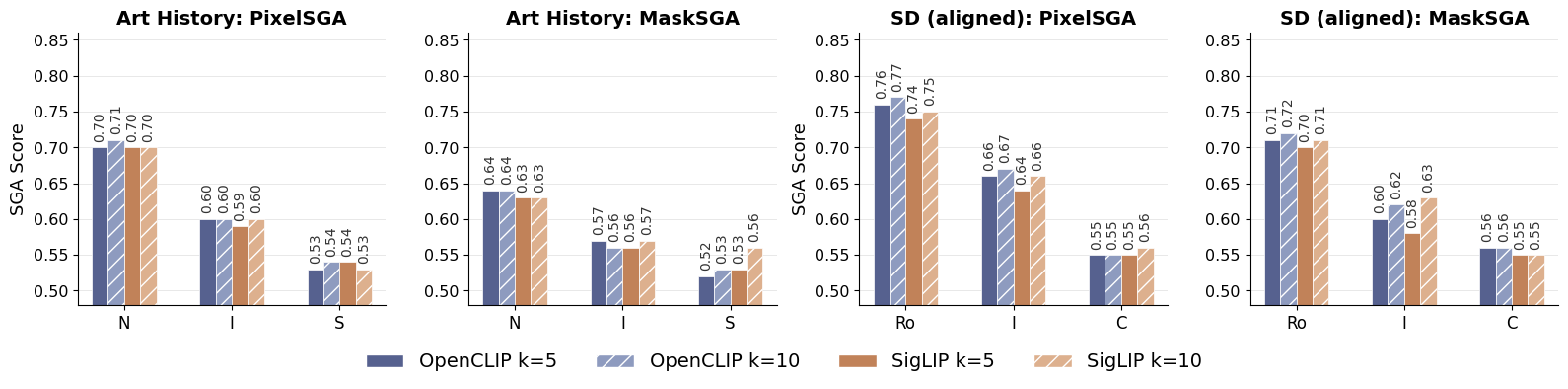}
    \caption{Robustness analysis of our metrics (PixelSGA and MaskSGA) to the choice of embedding model. Here we report results on both historical (Art History) and semantically aligned images generated with Stable Diffusion (SD). Style abbreviations: N (Neoclassicism), I (Impressionism), S (Symbolism), R (Rococo), and C (Cubism). For additional robustness analyses see Appendix~\ref{sec:apx:robustness}.}
    \label{fig:robustness_main}
\end{figure}

\paragraph{Robustness Analysis}
For each of the three presented experiment, we assess the robustness of the SGA measures by varying the embedding model (OpenCLIP vs.\ SigLIP \cite{zhai2023sigmoid}, \texttt{google/siglip-base-patch16-224}) and the number of neighbors ($k \in \{5,10\}$). We evaluate \textsc{PixelSGA} and \textsc{MaskSGA} on three representative styles (high-, medium-, and low-SGA values) and verify that their relative ordering remains consistent across configurations. In Figure \ref{fig:robustness_main}, we observe such behavior for Neoclassicism, Impressionism, and Symbolism in the art history setting, and for Rococo, Impressionism, and Cubism in the cross-domain experiment with \textit{SD}. Additional results, all validating the robustness of the SGA metrics, are provided in Appendix \ref{sec:apx:robustness}.

\section{Discussion}
\label{sec:discussion}

Our analysis shows that artistic styles in art history exhibit measurable gender artifacts, with Neoclassicism and Art Nouveau yielding the highest scores, reflecting their stylistic conventions (\textit{e.g.}, differentiated social and mythological gender roles in Neoclassicism \cite{neoclassicism_gender_representations} and stylized female representation in Art Nouveau \cite{artnouveau_universal_style}). In addition, gender information is encoded in style prompts, and is often amplified during generation. This suggests that stylistic conditioning in generative models is not socially neutral and that style replication is not strictly faithful. However, while gender artifacts are commonly used in the literature to analyze visual datasets \cite{hirota2025bias,meister2023gender}, their societal interpretation requires higher-level contextual analysis, which \textsc{StyleGender} supports, as explained next.

\paragraph{Dataset Usage} Given our foundational results on SGA measurements, we encourage the use of the \textsc{StyleGender} dataset to study high-level semantic factors such as skin tone, gendered objects or attributes, nudity, and broader contextual cues that may shape gender–style relationships and that are interpretable in terms of societal implications. The dataset indeed supports technical investigations of generative model behavior, while also enabling interpretations grounded in historical and cultural context. As a result, it has the potential of contributing to a range of fields, including AI research, computer vision, algorithmic fairness, cultural analytics, and digital art history.

\paragraph{Societal Implications}
Our findings show that even though style keywords do not explicitly encode gender semantics, they induce systematic shifts in gender-related visual patterns. This highlights risks associated with binary gender categorization, consistent with prior work on stereotype amplification in AI systems using categorical demographic labels \cite{doh2025position}. By measuring how T2I models amplify gender artifacts, we show how such binary annotations may reinforce socially constructed distinctions in image generation. We acknowledge that relying on fixed gender labels may inadvertently contribute to gender stereotyping. To mitigate potential misuse, we release the \textsc{StyleGender} dataset under a CC BY-NC-SA 4.0 license, restricting commercial use.

\paragraph{SGA limitations}
Although we conduct statistical sanity checks and robustness analyses, all SGA measures depend on classifier performance and inherit its limitations. Similarly, bounding-box detection and masking may be less reliable across more abstract styles due to differences in composition and figure placement \cite{ramos2025no,bengamra2024comprehensive}. 
While we propose two complementary measures of gender artifacts, alternative approaches may provide additional insights in future work investigations.

\paragraph{Dataset Limitations}
The use of \textsc{StyleGender} should be considered in light of several style-dependent limitations. First, the stylistic labels from WikiArt are treated as ground truth, but they are crowdsourced and may contain noise \cite{wasielewski2022beyond}. Moreover, the dataset presents a clear Western bias, with only one style (Japenese Ukiyo-e) not representing a Western artistic current. Second, in abstract and highly stylized styles, the perceived gender annotations in art historical images can be inherently ambiguous, and in T2I generation, style fidelity may vary due to uneven representation in model training data. Third, while manual corrections are applied, few residual mistakes are likely to exist in the dataset. In particular, the captions generated by Qwen2-VL (Fig.~\ref{fig:qwen}) may also reflect model-specific interpretive biases, including hallucinated or misidentified elements in ambiguous scenes. In this context, the 77-token limit of the CLIP text encoder constrains captioning, often requiring truncation that can omit relevant visual content; in rare cases, subjects were reintroduced manually from raw Qwen outputs. For highly abstract images where subjects were not detected, minimal gender annotations were added based on ground-truth labels (examples in Appendix~\ref{sec:apx:qwen}).

\paragraph{Computational Resources}
The experiments were conducted collaboratively across institutions using Google Colab with NVIDIA T4 GPUs (16\,GB VRAM) for preprocessing and analysis, and an institutional SLURM cluster with NVIDIA A6000 GPUs (48\,GB VRAM) for image generation. All stages were designed to run without multi-GPU parallelization. More details in Appendix \ref{sec:apx:computation}.

\section{Conclusion}

In this work, we introduce \textsc{StyleGender}, a dataset of around 74k images spanning 19 artistic styles, designed to study the intersection of gender representation and artistic style in both art history and T2I generation. As a primary analysis, we introduce two Set Gender Artifacts (SGA) metrics, namely \textsc{PixelSGA} and \textsc{MaskSGA}, to capture pixel-level and compositional gender-discriminative signals respectively. Our results show that gender artifacts vary systematically across historical styles, that style keywords alone modulate these patterns in T2I generation, and that generative models tend to amplify rather than reproduce historical gender artifacts. Together, these findings establish that artistic style is not socially neutral and it carries measurable gender-coded visual structure that generative models inherit and exacerbate. The \textsc{StyleGender} dataset is designed to support broader investigations beyond the analysis presented here, including gender-differentiated patterns in skin tone, nudity, pose, and object co-occurrence across styles.


\bibliographystyle{plain}
\bibliography{main}

\newpage

\appendix

\section{The \textsc{StyleGender} Dataset}
\label{sec:apx:dataset}

In this section we provide additional information on how the dataset \textsc{StyleGender} is constructed. 

\subsection{Art History}
\label{sec:apx:wiki}

This section first describes the WikiArt scraping and manual annotation procedure.

The art historical images in this study were downloaded from WikiArt using its advanced search. Images were filtered by stylistic and gender tags. Available gender annotations included:

\begin{itemize}
\item \textbf{Female:} \texttt{female-portraits}, \texttt{female-nude}, \texttt{female-slave}, \texttt{girls}, \texttt{queens}, \texttt{Mother}, \texttt{Woman warrior}
\item \textbf{Male:} \texttt{male-portraits}, \texttt{male-nude}, \texttt{Male}, \texttt{kings}, \texttt{boys}
\end{itemize}

Tags without a corresponding counterpart for the other gender indicate that no such label was available.

To ensure reliable gender classification, we retained only styles with a sufficient number of samples and reasonable class balance. Several styles were excluded due to very small sample sizes or extreme gender imbalance (\textit{e.g.}, Byzantine Art, Naturalism, Tonalism, Feminist Art, New Medievalism, Transautomatism, Magic Realism, Art Deco, and others). In addition, closely related subcategories (Early Renaissance, High Renaissance, Mannerism, and Northern Renaissance) were merged into a single \textit{Renaissance} category.

The downloaded images were manually reviewed to verify that the depicted subjects matched the ground-truth gender labels in terms of perceived gender. For highly abstract or stylized artworks, however, gender could not always be reliably inferred from the visual content alone. In such cases, the annotation was primarily guided by the title of the painting; when the title provided no relevant information, the original label was retained. Overall, the WikiArt tags were correct in the vast majority of cases, and only a small fraction of images needed to be excluded from the study because of incorrect gender label. In addition, pictures of sculptures or other art forms were also deleted from the dataset, keeping only paintings, drawings and digital art.

\subsection{Text-to-Image Generation}
This section details the generation of images with two T2I models using structured prompts.

\label{sec:apx:prompts}

Images were generated using the following prompt template:

\begin{quote}
\textit{``A painting of \textlangle gender\textrangle, \textlangle viewpoint\textrangle, \textlangle pose\textrangle{} [\textlangle modifier\textrangle], in the artistic style known as \textlangle style\textrangle''}
\end{quote}

\noindent Each variable took the following values:

\begin{itemize}
    \item \textbf{\textlangle gender\textrangle} (8 values): \textit{woman, man, female character, male character, female figure, male figure, human female, human male}
    \item \textbf{\textlangle viewpoint\textrangle} (5 values): \textit{full-body, half-body, torso-up, head-and-shoulders, close-up portrait}
    \item \textbf{\textlangle pose\textrangle} (5 values): \textit{facing forward, three-quarter view, profile view, upright neutral posture, arms relaxed at sides}
    \item \textbf{\textlangle modifier\textrangle} -- optional secondary pose detail (2 values, or absent): \textit{head slightly tilted, hands loosely clasped}
    \item \textbf{\textlangle style\textrangle} -- artistic style name, varied across generations
\end{itemize}

\noindent This yielded 500 unique prompt templates, distributed across framing groups: \textit{full-body} ($n=120$), \textit{close-up portrait} ($n=200$), and \textit{half-body}, \textit{torso-up}, and \textit{head-and-shoulders} ($n=60$ each).

\subsection{From Art History to Text-to-Image Generation}
\label{sec:apx:qwen}

This section details the caption generation and correction pipeline used to produce semantically aligned prompts for the historical image dataset, and provides qualitative examples of the resulting image generations.

\paragraph{Gender correction.}
Qwen2-VL-2B-Instruct occasionally describes human subjects using gender-neutral terms (e.g., \textit{figure}, \textit{person}, \textit{individual}) even when the gender of the depicted subject is visually unambiguous. To enforce gender consistency with the source artwork, we apply a post-processing step that replaces neutral terms with explicit gender labels derived from manual annotation of the original paintings. 
Figure~\ref{fig:apx:subject_Correction} illustrates this correction: the raw caption refers to the subject as \textit{figure} or \textit{person}, while the cleaned caption substitutes the correct gender label (\textit{woman} or \textit{man}), ensuring that the generative models receive unambiguous subject descriptions.
\begin{figure}[h]
    \centering
    \includegraphics[width=\linewidth]{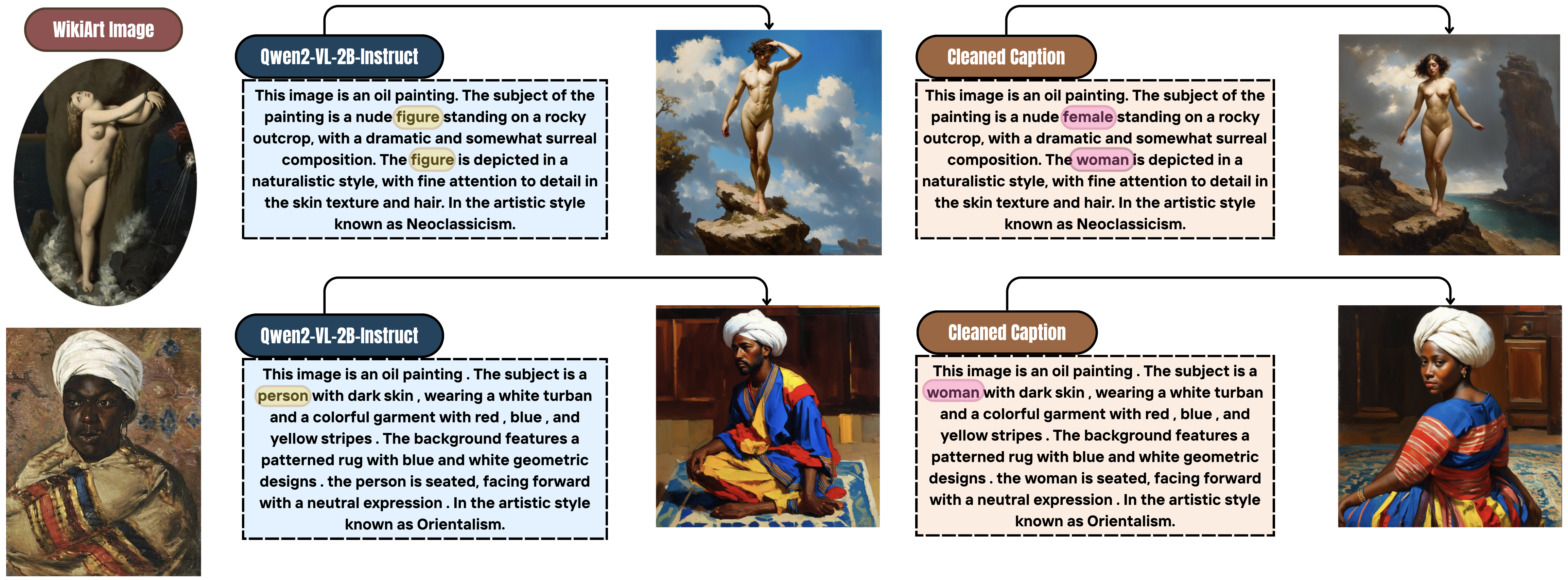}
    \caption{Examples of gender correction applied to Qwen2-VL-generated captions. For each painting, the raw caption (left) uses a gender-neutral term to refer to the subject (\textit{figure}, \textit{person}), while the cleaned caption (right) replaces it with an explicit gender label consistent with the source artwork. Corrected terms are highlighted in color. Paintings depicted (from top-to-bottom): "Angelica in Chains"
by Jean Auguste Dominique Ingres and "Portrait of a North African Lady" by Cesare Biseo.}
    \label{fig:gender_correction}
\end{figure}
\paragraph{Manual subject correction.}
In a small number of cases, Qwen2-VL failed to identify the primary human subject of the painting altogether, typically when the figure is partially occluded, located in the background, or compositionally marginal. In these cases, the model described the scene (landscape, objects, setting) without mentioning any person. Such captions were manually corrected to reintroduce an explicit subject description, preserving the remaining scene elements generated by the model. Figure~\ref{fig:apx:subject_Correction} shows a representative example: the raw Qwen2-VL caption describes the landscape in detail 
while omitting the human figures visible in the original; the corrected caption reintroduces a subject description while retaining the stylistic and compositional elements of the original output.
\begin{figure}[h]
    \centering
    \includegraphics[width=\linewidth]{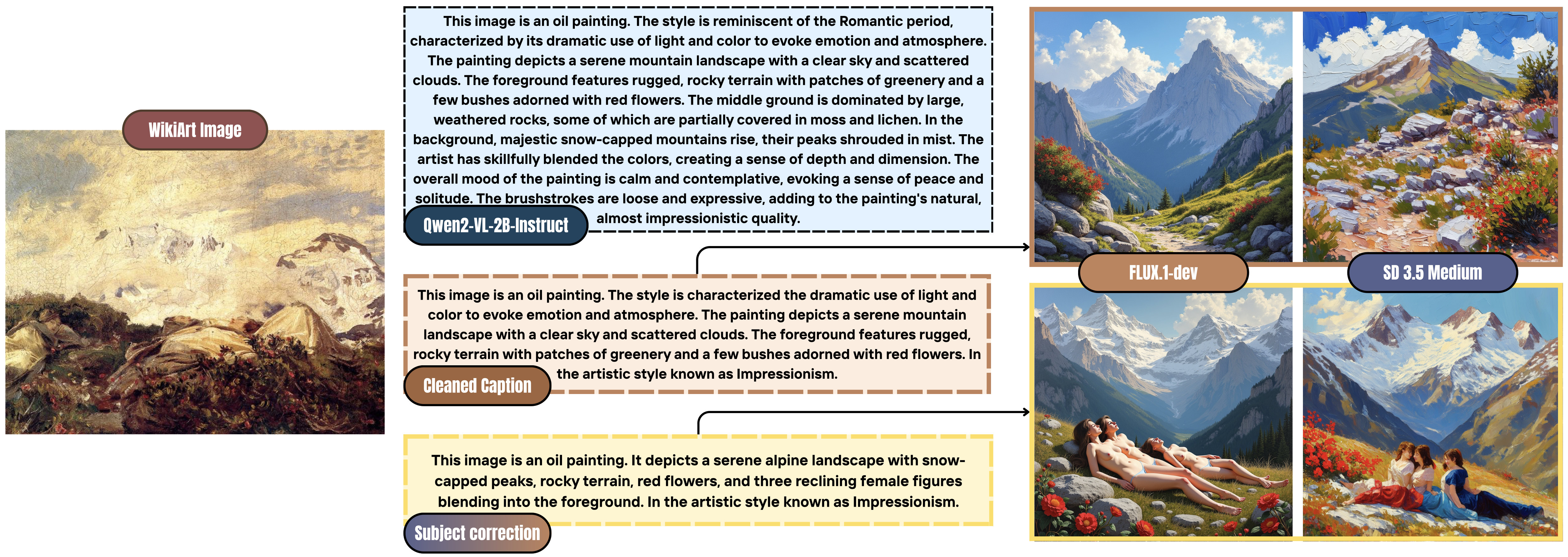}
    \caption{Example of manual subject correction. The raw Qwen2-VL caption (top) describes the landscape and atmosphere of the painting while omitting the human figures present in the scene. The corrected caption (bottom) reintroduces an  explicit subject description. The final corrected prompt used for generation is shown in the \textit{Subject correction} box. Painting depicted: "Princess Nouronihar" by John Singer Sargent.}
    \label{fig:apx:subject_Correction}
\end{figure}

\begin{figure}
    \centering
    \includegraphics[width=\linewidth]{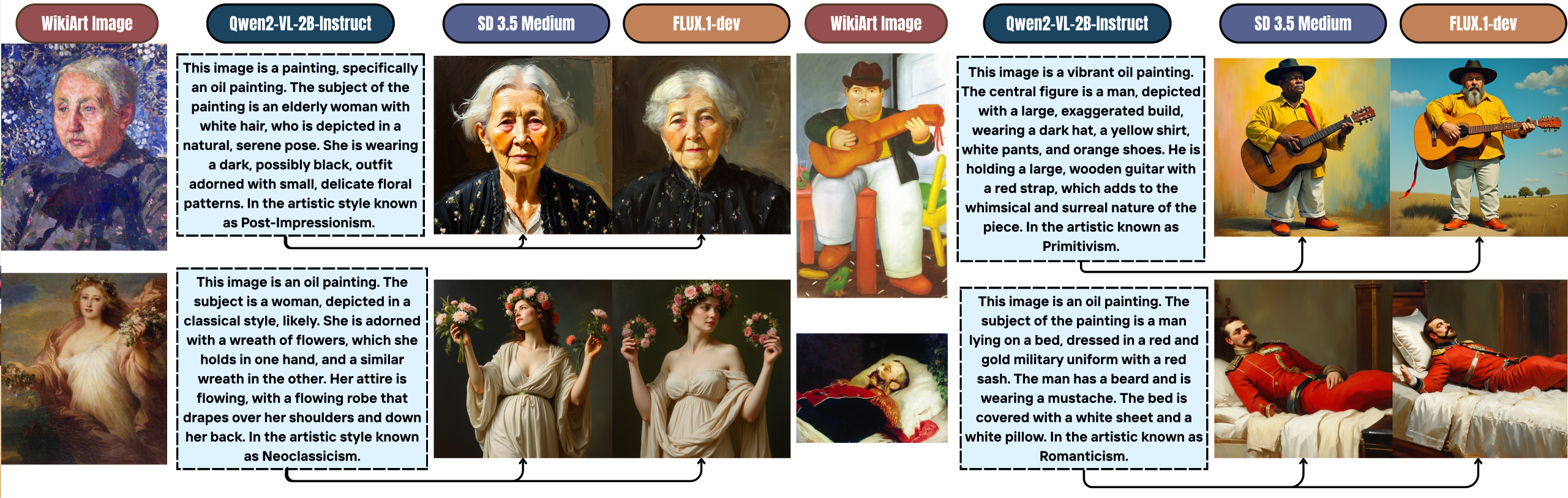}
    \caption{Qualitative examples of the captioning and generation pipeline. Each row shows a WikiArt source image, the corresponding Qwen2-VL cleaned caption, and images generated by \textit{SD~3.5~Medium} and \textit{FLUX.1-dev} conditioned on the cleaned caption. Paintings depicted: "Portrait of Madame Monnon, the Artist s Mother in Law" by Theo van Rysselberghe (top left), "The Spring" by Franz Xaver Winterhalter (bottom left), "Man with a guitar" by Fernando Botero (top right),  "Emperor Alexander II on his deathbed" by Konstantin Makovsky (bottom right).}
    \label{fig:examples_rq3_appen}
\end{figure}

Figure~\ref{fig:examples_rq3_appen} presents illustrative examples across multiple artistic styles and subject types, showing the original WikiArt image alongside the Qwen2-VL-cleaned caption and the images generated by \textit{SD~3.5~Medium} and \textit{FLUX.1-dev} conditioned on it. These examples illustrate that the captioning and correction pipeline produces prompts that faithfully preserve the semantic content of the source paintings, subject identity, pose, clothing, and scene composition, while allowing the generative model to reinterpret the image in the target artistic style

\section{Implementation of the Metrics}
In this section, we provide details on the implementation of the PixelSGA and MaskSGA measurements. 

\subsection{Bounding Boxes}
\label{sec:apx:bb}

Here, we describe the manual annotation of bounding boxes around people in the images, which are used for the computation of \textsc{MaskSGA} in our experiments.

Unlike in traditional object detection, where the aim is to obtain a single bounding box that well-captures the object, our aim is to ensure we mask each person in the image as best as possible, without overmasking the image.
We generate bounding boxes with DINO, transformer-based end-to-end detection model \cite{zhang2023dino}.\footnote{We also experimented with YOLO11, YOLO26 \cite{yolo_ultralytics}, and Facebook's DE:TR \cite{carion2020end}, but found them not to work as well on our art history dataset.}    
DINO gives multiple boxes $b_i$ with confidence scores $c_i \in [0,1]$.
Confidence scores of bounding boxes vary widely between styles and individual images, but we know there is at least one person per image.
In contrast to some other approaches, DINO does not use non-maximum suppression (NMS), where overlapping boxes are partly removed to avoid detecting objects multiple times.
We tested NMS (followed by keeping the top $k$ boxes) against a simpler approach that retains each box $b_i$ whose confidence $c_i$ is \textit{either} above a global threshold $t \in [0,1]$ \textit{or} within a relative margin $m \in [0,1]$ of the highest confidence $c_{\max}$ in that image:
\begin{equation}
    \{b_i \ | \ c_i \geq t\ \vee\ c_i \geq c_{\max} \cdot (1-m) \}. \label{eq:bb_mt}
\end{equation}
We manually annotated 40 randomly drawn images per style (in the art historical case) with boxes and calculated the overlap between the generated boxes and manual annotations. We display the results in Table~\ref{tab:bb_strategies}.
As we need to mask all human subjects to ensure that gender information in the subjects themselves is erased, recall is more important than precision for our task.
Based on this and its consistency across styles, we chose (\ref{eq:bb_mt}) with $t=m=0.3$ for selecting the set of bounding boxes per image.

\begin{table}[ht]
\centering
\caption{Different strategies for selecting bounding boxes: All boxes, the box with highest confidence, versions of NMS, or the strategy of (\ref{eq:bb_mt}) with different choices of $t$ and $m$. We selected t$=$0.3, m$=$0.3 for high recall and reasonable precision.}
\label{tab:bb_strategies}
    \small
\begin{tabular}{lcccc}
    \hline
    \textbf{Strategy} & \textbf{Precision} & \textbf{Recall} & \textbf{F1} & \textbf{Mean IoU} \\
    \hline
    all boxes & 0.284 & 0.929 & 0.435 & 0.961 \\
    top-1       & 0.854 & 0.633 & 0.727 & 0.923 \\
    NMS k$=$3    & 0.706 & 0.743 & 0.724 & 0.923 \\
    NMS k$=$5 & 0.594 & 0.828 & 0.692 & 0.925 \\
    t$=$0.3, m$=$0.1  & 0.745 & 0.774 & 0.759 & 0.933 \\
    t$=$0.3, m$=$0.3  & 0.633 & 0.800 & 0.707 & 0.936 \\
    t$=$0.3, m$=$0.5  & 0.453 & 0.847 & 0.590 & 0.945 \\
    \hline
\end{tabular}%
\end{table}

\subsection{The \textit{monotonicity score}}
\label{sec:apx:mono}

\begin{figure}[h!]
  \centering
  \includegraphics[width=\textwidth]{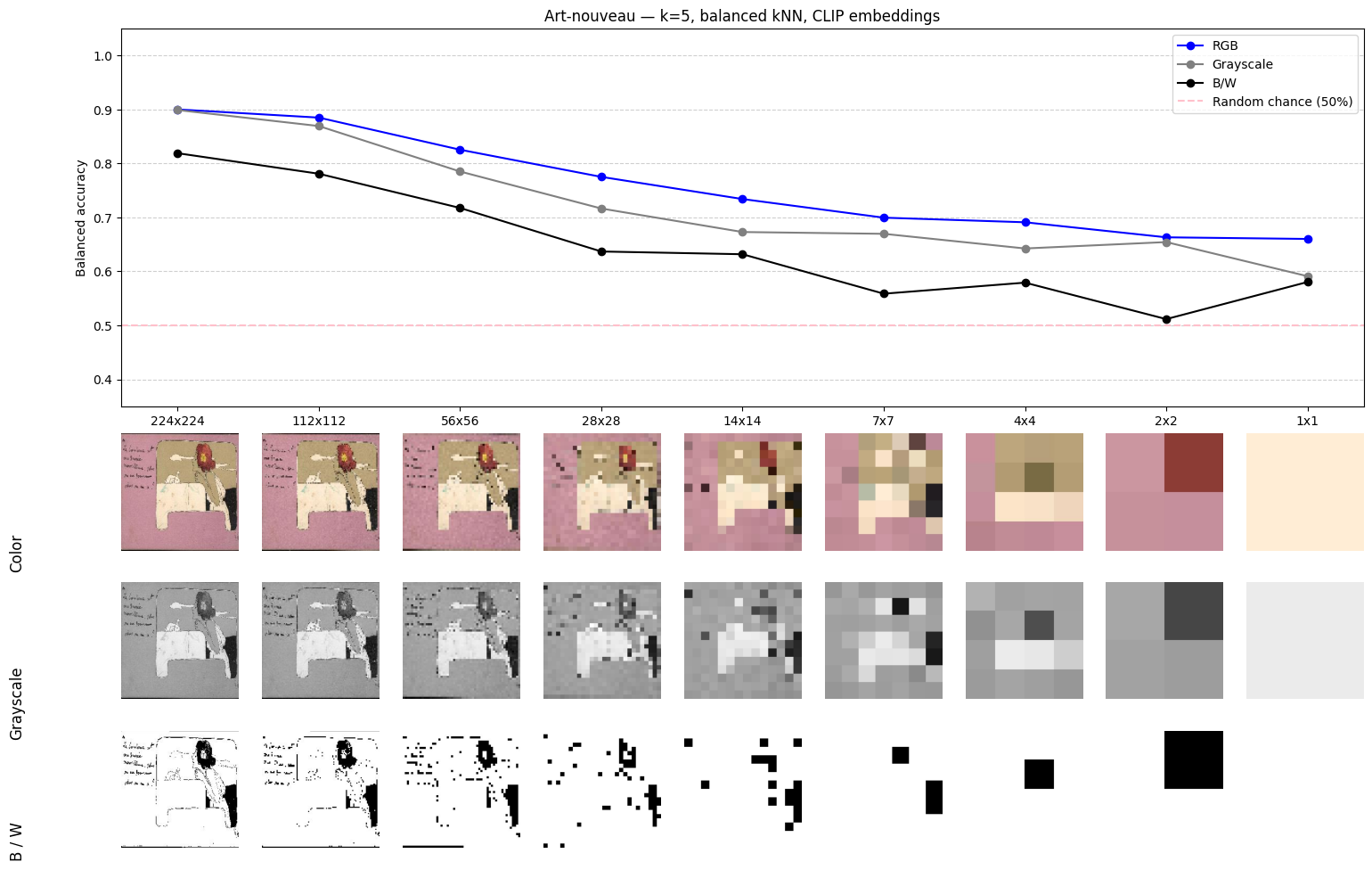}
  \caption{Classifier performance on Art Nouveau art historical images at different resolutions and different chromatic variations.
  }
  \label{fig:artnouveau}
\end{figure}

\begin{figure}[h!]
  \centering
  \includegraphics[width=\textwidth]{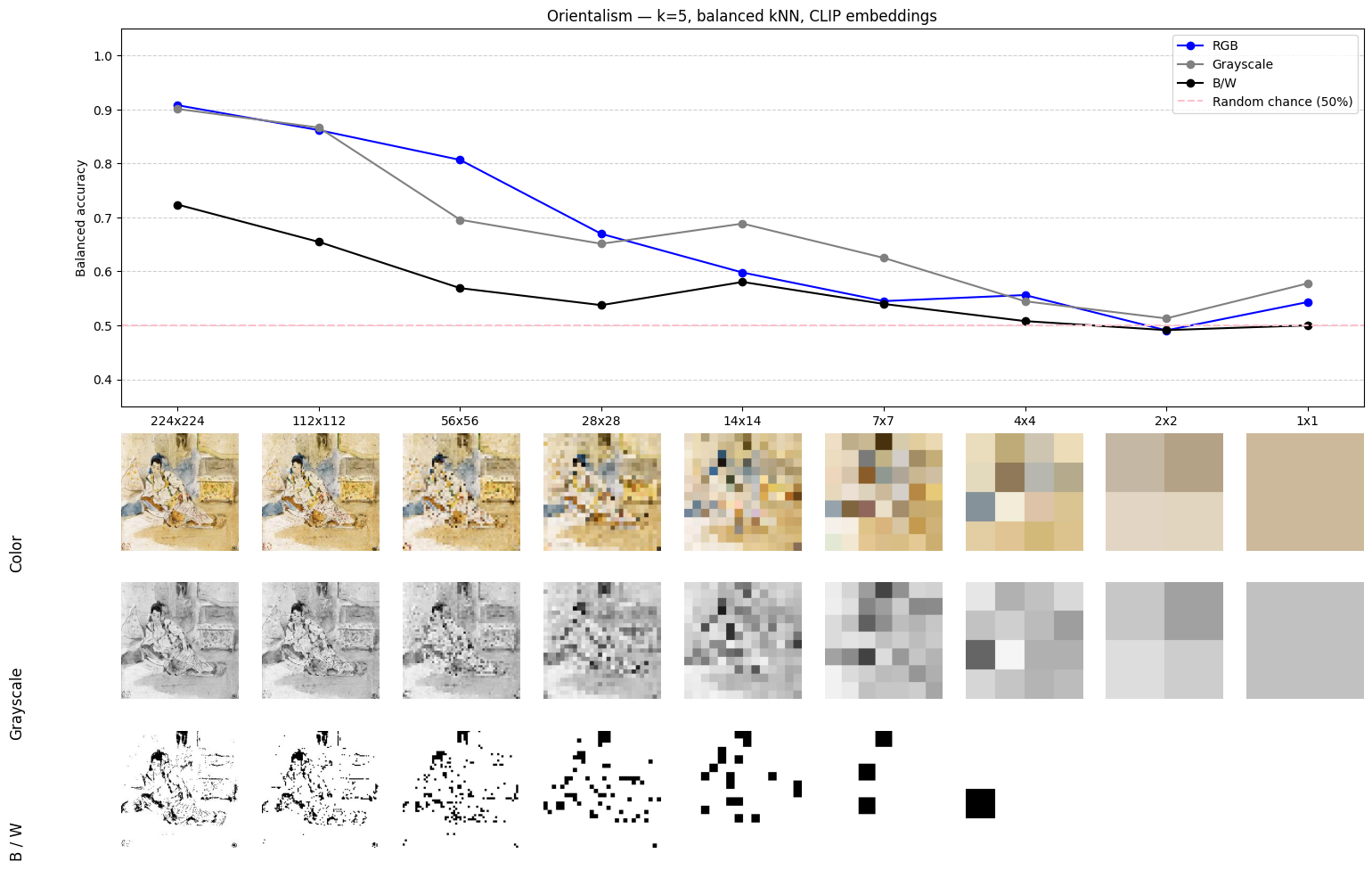}
  \caption{Classifier performance on Orientalism art historical images at different resolutions and different chromatic variations.
  }
  \label{fig:orientalism}
\end{figure}

In this section, we discuss a validation of our CLIP- and kNN-based classifier that is used for the metrics.
A natural worry may be that CLIP is trained on full images, so it is less clear whether our kNN classifier on embeddings works as intended on transformed images.
We perform a sanity check as follows. By downsampling to lower and lower resolutions, the information content of the images is more and more reduced.
This means that the accuracy of our classifier should decrease roughly monotonically as we downsample further.
We illustrate this behavior in Figures \ref{fig:artnouveau} and \ref{fig:orientalism}.
To capture this quantitatively, we compute Spearman's rank correlation coefficient $\rho$ between resolution and balanced accuracy.
We do this for all experiments, styles, and encodings, and report the average correlation coefficient per experiment as a monotonicity score in Section~\ref{sec:experiments}.
In line with best practices, we average over the z-statistics of the coefficients $\rho$ rather than the coefficients themselves, and then transform the result back:
\begin{equation*}
    \rho_{e} = \mathrm{tanh} \left(\frac{1}{|\mathcal{T}||\mathcal{S}|} \sum_{\substack{t \in \mathcal{T} \\ s \in \mathcal{S}}} \mathrm{tanh}^{-1}(\rho_{e,t,s}) \right)
\end{equation*}
where $\rho_{t,s,e}$ is Spearman's rank correlation coefficient for experiment $e$, chromatic encoding $t$, and style $s$.

\section{Additional Results}

\subsection{Gender Artifacts in Art History}
\label{sec:apx:wiki_results}

\begin{table}[h]
\centering
\tiny 
\setlength{\tabcolsep}{2pt} 
\renewcommand{\arraystretch}{0.8} 
\caption{Full table of results on the art history case for the spatial and chromatic manipulations. The highlighted values are those contributing to the \textsc{PixelSGA} for each style. "G" stands for "grayscale", "B/W" refers to binary chromatic variation; "P. Impressionism" stands for "Post-Impressionism", "C. Realism" stands for "Contemproary Realism".}
\resizebox{\textwidth}{!}{%
\begin{tabular}{l l ccccccccc l l ccccccccc}
\toprule
\textbf{Style} & \textbf{Mode} & 224 & 112 & 56 & 28 & 14 & 7 & 4 & 2 & 1 &
\textbf{Style} & \textbf{Mode} & 224 & 112 & 56 & 28 & 14 & 7 & 4 & 2 & 1 \\
\midrule
\multirow{3}{*}{Academicism} & RGB       &  .93   &  .92   &  .87   &  .82   &   \textbf{.73}  &  \textbf{.71}   &   \textbf{.67} &   \textbf{.61}  &   .57   &
\multirow{3}{*}{Art Nouveau} & RGB       &  .90   &  .88   &  .83   &   .78  &   \textbf{.73}  &  \textbf{.70}   & \textbf{.69}  &   \textbf{.66}  &  \textbf{.66}  \\
    & G& .92     &  .92   & .89    &  .77   &  .69   &   .68   &  .64   &  .59   &   .55  &
  & G &  .90   &   .87  &  .79   &  .72   &   .67   &  .67   &  .64   &  .65   &  .59   \\
& B/W       &  .84   &   .76  &  .69   &  .71   &     .69 &  .64   &   .64  &    .59 &   \textbf{.58}  &
& B/W       &   .82  &  .78   &  .72   &  .64   &  .63   &  .56   &  .58   &   .51  &   .58  \\
\midrule
\multirow{3}{*}{Baroque}     & RGB       &   .93  &   .92  &  .84   &  .69   &  \textbf{.62}   &   .53  &  .54   &   \textbf{.54}  &  \textbf{.57}   &
\multirow{3}{*}{C. Realism} & RGB & .84    &  .79   &   .79  &  .66   &  \textbf{.68}   &  .57   &  \textbf{.63}   &  \textbf{.58}   &   \textbf{.60}  \\
 & G&   .92  &  .89   &  .80   &   .66  &  .60   &   .56  &   .55  &   .47  &   .51  &
 & G &  .83   &   .78  &  .78   &  .66   &  .59   &   \textbf{.61}  &    .51 &  .50   &  .52   \\
& B/W       &  .76   &  .69   & .63    &  .57   &  .57   &   \textbf{.57}  &  \textbf{.59}   &   .50  &   .40  &
& B/W       &   .73  &  .63   & .62    &  .62   &   .55  &   .57  &  .51   &   .49  &   .50  \\
\midrule
\multirow{3}{*}{Cubism}      & RGB       &   .78  &   .71  &   .63  & .59    &  \textbf{.57}   &   \textbf{.59}  &  \textbf{.54}   &  \textbf{.54}   &  .49   &
\multirow{3}{*}{Expressionism} & RGB       &  .88   &  .85   &  .77   &  .67   &  \textbf{.59}   &  \textbf{.56}   &  \textbf{.54}   &   \textbf{.53}  &   .51  \\
& G &   .71  &  .70   &   .62  &   .48  &  .47   &   .46  &  .50   &  .52   &  \textbf{.50}   &
& G&   .85  &  .83   &   .73  &   .61  &   .55  &  .54   &   .54  &  .51   &  .51   \\
& B/W       &   .72  &   .68  &   .58  &   .50  &  .49   &   .44  &   .45  &  .53   &  .50   &
& B/W       &  .77   &   .70  &   .60  &   .55  &  .52   &   .52  &   .54  &  .51   &  \textbf{.52}   \\
\midrule
\multirow{3}{*}{Impressionism} & RGB     &  .93   &  .91   &    .84 &   .73  &  \textbf{.64}   &  \textbf{.61}   &  \textbf{.58}   &   \textbf{.57}  & \textbf{.58}    &
\multirow{3}{*}{Neoclassicism}      & RGB       &  .97   &    .96 &  .92   &   .84  &  .79   &   \textbf{.70}  &    .68 &  \textbf{.64}   &  \textbf{.65}   \\
& G &  .91   &   .88  &   .78  &   .67  &    .59 &   .58  &  .56   &  .57   &  .54   &
& G &  .97   &   .96  &   .91  &  .79   &    .74 &  .69   &   \textbf{.69}  &  .62   &  .56   \\
& B/W       &  .77   &   .68  &  .60   &  .58   &   .58  &  .56   &   .54  &   .57  &   .47  &
& B/W  &  .89   &  .83   &  .75   &   .75  &    .70 &  .65   &   .67  &  .59  &  .52   \\
\midrule
\multirow{3}{*}{Orientalism} & RGB     &  .91   &  .87   &    .81 &  .67   &  .60   &   .55  &   \textbf{.56}  &    .49 &  .54   &
\multirow{3}{*}{Pop-Art}   & RGB     &  .88   &   .85  &  .83   &   .74  &  \textbf{.69}   &  .61   &   \textbf{.55}  &  \textbf{.61}   & \textbf{.55}    \\
& G &  .90   &  .87   &  .70   &   .65  & \textbf{.69} &   \textbf{.63} &   .54  &  \textbf{.51}   &  \textbf{.58}   &    
& G &  .89   &  .86   &  .74   &  .67   & .64    &  \textbf{.64}   & .55    &  .50   &  .46   \\
& B/W       &   .72  & .65    &  .57   &  .54   &  .58   &  .54   &  .51   &  .49   &   .50  &
& B/W       &   .83  &  .80   &  .63   & .61 &  .54   &  .51   &  .48   & .48    &  .50      \\
\midrule
\multirow{3}{*}{P. Impressionism} & RGB      & .91    &   .88  & .79    &  .66   &  \textbf{.57}   & .54 &  .51   &  .52   &  .56   &
\multirow{3}{*}{Primitivism}    & RGB      & .84    &   .84  & .81    &  .66   &  .54   & .54 &  \textbf{.61}   &  \textbf{.55}   &  \textbf{.56}   \\
& G &  .91   &  .87  &   .74  &   .57  &   .55  &  \textbf{.55} & .49   &  .51   &  \textbf{.57}   & 
& G &  .83   &  .83   &   .73  &   .59  &   \textbf{.60}  &  \textbf{.60} & .57   &  .52   &  .55   \\
& B/W       &  .78   &    .71 &  .64   &  .56   &  .56   &  .54   &  \textbf{.52}   &  \textbf{.55}   &  .50   &
& B/W       &  .80   &    .70 &  .62   &  .55   &  .56   &  .53   &  .55   &  .54   &  .50   \\
\midrule
\multirow{3}{*}{Realism}       & RGB      &  .94   &  .92   &   .84  &  .73   &   \textbf{.65}  &  \textbf{.59}   &  \textbf{.57}   &  .56   &  .56   &
\multirow{3}{*}{Renaissance} & RGB  & .92    &  .90   &  .81   &  .74   &  \textbf{.68}   &  .60   &  .54   &  .53   & \textbf{.56}    \\
& G &  .94   &   .91  &   .82  &   .68  &  .61   &  .58   &  .57   &  \textbf{.57}   &  .53   &
& G &  .92   &  .89   &   .79  &  .68   &  .65   &  \textbf{.61}   &  .56   &  \textbf{.54}   &  .56   \\
& B/W       &   .81  &  .74   &  .67   &  .61   &  .60   &  .59   &   .56  &  .54   &  \textbf{.59}   &
& B/W       &   .83  &  .76   &  .69   &  .64   &  .59   &  .60   &  \textbf{.57}   &  .52   &  .48   \\
\midrule
\multirow{3}{*}{Rococo}        & RGB      &  .96   &  .95   &  .93   &  .84   &  \textbf{.73}   &  .62   &  .61   &  .52   &  \textbf{.59}   &
\multirow{3}{*}{Romanticism}    & RGB      &  .95   &  .94   &   .88  & .78   &  \textbf{.71}   &  \textbf{.66}   &  \textbf{.63}   &  \textbf{.60}   &  .56   \\
& G &  .96   &  .95   & .91    &  .78   &  .67   &  \textbf{.63}   &  \textbf{.62}   &  \textbf{.60}   &  .59   &
& G &  .95   &  .93   & .86    &  .70   &  .65   &  .61   &  .60   & .59    &  .53   \\
& B/W       &  .87   &  .82   &  .75   &  .67   &  .65   &  .65   &  .60   &  .58   &  .55   & 
& B/W       &   .83  &  .78   &  .68   &  .65   &  .63   &  .57   &  .56   &  .56   &  \textbf{.62}   \\
\midrule
\multirow{3}{*}{Surrealism}    & RGB      &  .82   &  .79   &  .73   &  .68   &  \textbf{.65}   &  \textbf{.60}   &  .50   &  .54   &  \textbf{.53}   &
\multirow{3}{*}{Symbolism}     & RGB      &  .83   &  .77   &  .68   &  .58   &  \textbf{.64}   &  \textbf{.55}   &  .49   &  .51   &  \textbf{.52}   \\
& G &   .81  &  .77   &   .72 &  .61   &  .57   &  .56   &  \textbf{.55}   &  \textbf{.55}   &  .52   &
& G &  .80   &  .71   &   .65  &  .56   &  .51   &  .48   &  .49   & \textbf{.53}    &  .45   \\
& B/W       &  .71   &  .71   & .61    &  .56   &  .50   &  .53   & .52    & .46    &  .50   &
& B/W       &   .63  &   .59  &  .59   &  .52   & .54    &  .46   &  \textbf{.51}   &  .50   & .51    \\
\midrule
\multirow{3}{*}{Ukiyo-e}       & RGB      &  .88   & .86    &  .73   &  .71   &  \textbf{.72}   &  \textbf{.70}   &  \textbf{.62}   &  \textbf{.55}   &  \textbf{.53}   &
&          &     &     &     &     &     &     &     &     &     \\
& G &   .87  &  .77   &  .71   &  .71   &  .68   &  .62   &  .56   &  .55   & .51    &
&          &     &     &     &     &     &     &     &     &     \\
& B/W       & .82    & .73    &  .69   &  .66   &  .68   &  .60   &  .59   &  .52   &  .50   &
&          &     &     &     &     &     &     &     &     &     \\
\bottomrule
\end{tabular}%
}
\label{tab:rq1_pixelsga}
\end{table}

In Table \ref{tab:rq1_pixelsga}, we report additional results for the computation of \textsc{PixelSGA}. For completeness, the table also includes higher resolutions that are not meaningful for the metric (namely 224$\times$224, 112$\times$112, 56$\times$56, and 28$\times$28). As explained in the main paper, we restrict our analysis to resolutions below 16$\times$16, since higher resolutions may preserve trivial gender cues. In particular, more realistic artistic styles could retain easily recognizable gender features at higher resolutions, artificially inflating their \textsc{PixelSGA} values compared to more abstract or stylized artworks.

In the table, we highlight in bold the values contributing to \textsc{PixelSGA} across the three chromatic variants (RGB, grayscale, and binary). In many cases, the RGB version contains the most gender-discriminative information. However, this is not always the case. Figure \ref{fig:artnouveau} illustrates the progression of gender classifier performance for Art Nouveau images, where the contributing values consistently correspond to the RGB variant. In contrast, for the Orientalism example shown in Figure \ref{fig:orientalism}, the grayscale representation dominates in most resolutions used for the \textsc{PixelSGA}, with the exception of 4$\times$4, where RGB is only marginally higher.

\begin{table}[h]
\centering
\caption{Full table of results on the art history case for the occlusion manipulations. All these values contribute to the \textsc{MaskSGA} for each style. "P. Impressionism" and "C. Realism" respectively stand for "Post-Impressionism" and "Contemporary Realism".}
\setlength{\tabcolsep}{8pt} 
\begin{tabular}{lcc|lcc}
\toprule
\textbf{Style} & \textbf{Mask} & \textbf{MaskNoBg} & \textbf{Style} & \textbf{Mask} & \textbf{MaskNoBg} \\
\midrule
Academicism &  .66   &  .54  & Art Nouveau &  .73   &  .53   \\
Baroque &  .54   &  .51   & C. Realism &  .70   &   .58  \\
Cubism &  .64   &  .48   & Expressionism &  .57   &  .55   \\
Impressionism &  .60   &  .53   & Neoclassicism &  .65   &  .62   \\
Orientalism &   .65  &  .57   & Pop-Art &  .63   &   .63  \\
P. Impressionism &  .59   &  .53   & Primitivism &  .60   &   .54  \\
Realism &  .60   &  .50   & Renaissance &  .54   &  .53   \\
Rococo &  .57   &   .52  & Romanticism &   .59  &  .52   \\
Surrealism &   .63  &  .54   & Symbolism &  .55   &  .49   \\
Ukiyo-e &  .65   &  .55   &  &     &     \\
\bottomrule
\end{tabular}%
\label{tab:rq1_masksga}
\end{table}

In addition, in Table \ref{tab:rq1_masksga}, we report the results of the classifier for each style when masking the subject (Mask) and when masking also the background with a different color (MaskNoBg). All these values contribute to the computation of \textsc{MaskSGA} for each style.

\subsection{Gender Artifacts in Text-to-Image Generation}
\label{sec:apx:rq2_results}

In Table \ref{tab:rq2_pixelsga_sd} and Table \ref{tab:rq2_pixelsga_flux}, we report additional results for the computation of \textsc{PixelSGA} on these images, respectively for Stable Diffusion (SD) and Flux. In these tables, we highlight in bold the balanced accuracy values contributing to \textsc{PixelSGA} across the three chromatic variants (RGB, grayscale, and binary), as defined in (\ref{eq:pixelSGA}). In most cases, these correspond to RGB variations. In addition, in Table \ref{tab:rq2_masksga_sd} and  Table \ref{tab:rq2_masksga_flux} (for SD and Flux, respectively), we report the balanced accuracy results of the classifier for each style when masking the subject (Mask) and when masking also the background with a different color (MaskNoBg). All these values contribute to the computation of \textsc{MaskSGA} for each style, as defined in (\ref{eq:maskSGA}).

\begin{table}[h]
\centering
\tiny 
\setlength{\tabcolsep}{2pt} 
\renewcommand{\arraystretch}{0.8} 
\caption{Full table of results on the Stable Diffusion 3.5 Medium (SD) case for the spatial and chromatic manipulations. The highlighted values are those contributing to the \textsc{PixelSGA} for each style. "G" stands for "grayscale", "B/W" refers to binary chromatic variation; "P. Impressionism" stands for "Post-Impressionism", "C. Realism" stands for "Contemporary Realism".}
\resizebox{\textwidth}{!}{%
\begin{tabular}{l l ccccccccc l l ccccccccc}
\toprule
\textbf{Style} & \textbf{Mode} & 224 & 112 & 56 & 28 & 14 & 7 & 4 & 2 & 1 &
\textbf{Style} & \textbf{Mode} & 224 & 112 & 56 & 28 & 14 & 7 & 4 & 2 & 1 \\
\midrule
\multirow{3}{*}{Academicism} & RGB & 1.00 & 1.00 & 1.00 & .98 & \textbf{.89} & \textbf{.71} & \textbf{.68} & .56 & \textbf{.61} & \multirow{3}{*}{Art Nouveau} & RGB & 1.00 & 1.00 & 1.00 & .98 & \textbf{.93} & \textbf{.78} & \textbf{.69} & \textbf{.64} & \textbf{.67} \\
 & G & 1.00 & 1.00 & 1.00 & .94 & .78 & .67 & .59 & \textbf{.59} & .56 &  & G & 1.00 & 1.00 & 1.00 & .93 & .74 & .68 & .61 & .60 & .56 \\
 & B/W & .99 & .97 & .91 & .72 & .64 & .59 & .57 & .46 & .44 &  & B/W & .99 & .99 & .90 & .74 & .64 & .58 & .57 & .58 & .44 \\
\midrule
\multirow{3}{*}{Baroque} & RGB & 1.00 & 1.00 & 1.00 & .99 & \textbf{.92} & \textbf{.78} & \textbf{.74} & \textbf{.67} & \textbf{.71} & \multirow{3}{*}{C. Realism} & RGB & 1.00 & 1.00 & 1.00 & .99 & \textbf{.93} & \textbf{.79} & \textbf{.73} & \textbf{.66} & \textbf{.59} \\
 & G & 1.00 & 1.00 & 1.00 & .94 & .80 & .70 & .67 & .61 & .67 &  & G & 1.00 & 1.00 & 1.00 & .96 & .78 & .62 & .60 & .53 & .59 \\
 & B/W & 1.00 & .99 & .94 & .79 & .68 & .67 & .55 & .49 & .49 &  & B/W & .99 & .98 & .92 & .75 & .65 & .54 & .61 & .53 & .50 \\
\midrule
\multirow{3}{*}{Cubism} & RGB & 1.00 & 1.00 & .99 & .96 & \textbf{.81} & \textbf{.64} & \textbf{.59} & \textbf{.55} & \textbf{.57} & \multirow{3}{*}{Expressionism} & RGB & 1.00 & 1.00 & 1.00 & .99 & \textbf{.89} & \textbf{.70} & \textbf{.66} & \textbf{.57} & \textbf{.60} \\
 & G & 1.00 & 1.00 & .99 & .93 & .69 & .59 & .53 & .50 & .54 &  & G & 1.00 & 1.00 & .99 & .92 & .80 & .60 & .57 & .48 & .55 \\
 & B/W & .99 & .98 & .88 & .73 & .60 & .57 & .56 & .49 & .37 &  & B/W & .99 & .97 & .88 & .70 & .63 & .56 & .52 & .46 & .44 \\
\midrule
\multirow{3}{*}{Impressionism} & RGB & 1.00 & 1.00 & 1.00 & .99 & \textbf{.94} & \textbf{.80} & \textbf{.74} & \textbf{.69} & \textbf{.64} & \multirow{3}{*}{Neoclassicism} & RGB & 1.00 & 1.00 & 1.00 & .98 & \textbf{.87} & \textbf{.72} & \textbf{.70} & \textbf{.61} & \textbf{.71} \\
 & G & 1.00 & 1.00 & .99 & .93 & .84 & .69 & .59 & .56 & .61 &  & G & 1.00 & 1.00 & 1.00 & .94 & .73 & .63 & .58 & .51 & .60 \\
 & B/W & .99 & .98 & .90 & .77 & .62 & .55 & .58 & .61 & .45 &  & B/W & .99 & .96 & .90 & .68 & .66 & .56 & .57 & .50 & .48 \\
\midrule
\multirow{3}{*}{Orientalism} & RGB & 1.00 & 1.00 & 1.00 & .99 & \textbf{.92} & \textbf{.76} & \textbf{.68} & .53 & \textbf{.65} & \multirow{3}{*}{Pop-Art} & RGB & 1.00 & 1.00 & 1.00 & .98 & \textbf{.86} & \textbf{.71} & \textbf{.61} & \textbf{.59} & .59 \\
 & G & 1.00 & 1.00 & .99 & .96 & .77 & .60 & .57 & \textbf{.57} & .59 &  & G & 1.00 & 1.00 & .99 & .88 & .69 & .58 & .55 & .54 & .52 \\
 & B/W & .99 & .98 & .82 & .68 & .62 & .58 & .58 & .50 & .60 &  & B/W & 1.00 & 1.00 & .93 & .76 & .64 & .54 & .52 & .56 & \textbf{.62} \\
\midrule
\multirow{3}{*}{P. Impressionism} & RGB & 1.00 & 1.00 & 1.00 & .99 & \textbf{.93} & \textbf{.80} & \textbf{.71} & \textbf{.61} & \textbf{.62} & \multirow{3}{*}{Primitivism} & RGB & 1.00 & 1.00 & 1.00 & .94 & \textbf{.79} & \textbf{.67} & \textbf{.62} & \textbf{.59} & .54 \\
 & G & 1.00 & 1.00 & 1.00 & .95 & .81 & .70 & .64 & .53 & .55 &  & G & 1.00 & 1.00 & .99 & .87 & .72 & .58 & .54 & .51 & .56 \\
 & B/W & .99 & .98 & .88 & .72 & .66 & .60 & .53 & .46 & .45 &  & B/W & .97 & .95 & .83 & .68 & .62 & .53 & .55 & .59 & \textbf{.62} \\
\midrule
\multirow{3}{*}{Realism} & RGB & 1.00 & 1.00 & 1.00 & .99 & \textbf{.92} & \textbf{.75} & \textbf{.71} & \textbf{.61} & \textbf{.65} & \multirow{3}{*}{Renaissance} & RGB & 1.00 & 1.00 & 1.00 & .99 & \textbf{.91} & \textbf{.79} & \textbf{.68} & \textbf{.61} & \textbf{.73} \\
 & G & 1.00 & 1.00 & 1.00 & .95 & .83 & .70 & .61 & .57 & .59 &  & G & 1.00 & 1.00 & 1.00 & .96 & .81 & .68 & .59 & .56 & .63 \\
 & B/W & .98 & .97 & .90 & .76 & .67 & .57 & .59 & .51 & .50 &  & B/W & 1.00 & .98 & .94 & .77 & .66 & .64 & .60 & .55 & .35 \\
\midrule
\multirow{3}{*}{Rococo} & RGB & 1.00 & 1.00 & 1.00 & .99 & \textbf{.94} & \textbf{.82} & \textbf{.78} & \textbf{.65} & \textbf{.74} & \multirow{3}{*}{Romanticism} & RGB & 1.00 & 1.00 & 1.00 & .99 & \textbf{.93} & \textbf{.77} & \textbf{.72} & \textbf{.68} & \textbf{.72} \\
 & G & 1.00 & 1.00 & 1.00 & .96 & .83 & .75 & .66 & .64 & .64 &  & G & 1.00 & 1.00 & .99 & .95 & .78 & .71 & .65 & .62 & .60 \\
 & B/W & 1.00 & .99 & .92 & .81 & .73 & .65 & .57 & .42 & .45 &  & B/W & .99 & .97 & .88 & .74 & .67 & .58 & .59 & .58 & .61 \\
\midrule
\multirow{3}{*}{Surrealism} & RGB & 1.00 & 1.00 & 1.00 & .97 & \textbf{.89} & \textbf{.70} & \textbf{.64} & .58 & \textbf{.65} & \multirow{3}{*}{Symbolism} & RGB & 1.00 & 1.00 & 1.00 & .98 & \textbf{.83} & \textbf{.73} & \textbf{.67} & .56 & \textbf{.64} \\
 & G & 1.00 & 1.00 & .99 & .91 & .78 & .68 & .59 & .57 & .57 &  & G & 1.00 & 1.00 & .99 & .92 & .73 & .63 & .59 & .62 & .57 \\
 & B/W & .99 & .97 & .88 & .73 & .61 & .54 & .54 & \textbf{.61} & .51 &  & B/W & .98 & .96 & .85 & .70 & .63 & .56 & .56 & \textbf{.64} & .38 \\
\midrule
\multirow{3}{*}{Ukiyo-e} & RGB & .98 & .98 & .94 & .92 & \textbf{.79} & \textbf{.70} & \textbf{.65} & \textbf{.61} & \textbf{.68} &  & &  &  &  &  &  &  &  &  &  \\
 & G & .99 & .99 & .94 & .81 & .69 & .66 & .59 & .55 & .63 &  &  &  &  &  &  &  &  &  &  &  \\
 & B/W & .98 & .96 & .89 & .72 & .65 & .56 & .56 & .61 & .61 &  & &  &  &  &  &  &  &  &  &  \\
\bottomrule
\end{tabular}%
}
\label{tab:rq2_pixelsga_sd}
\end{table}

\begin{table}[h]
\centering
\caption{Full table of results on the Stable Diffusion 3.5 Medium (SD) case with neutral prompts, considering the occlusion manipulations. All these values contribute to the \textsc{MaskSGA} for each style. "P. Impressionism" and "C. Realism" respectively stand for "Post-Impressionism" and "Contemporary Realism".}
\setlength{\tabcolsep}{8pt} 
\begin{tabular}{lcc|lcc}
\toprule
\textbf{Style} & \textbf{Mask} & \textbf{MaskNoBg} & \textbf{Style} & \textbf{Mask} & \textbf{MaskNoBg} \\
\midrule
Academicism &  .60   &  .56  & Art Nouveau &  .59   &  .58   \\
Baroque &  .67   &  .59   & C. Realism &  .64   &   .52  \\
Cubism &  .68   &  .50   & Expressionism &  .63   &  .57   \\
Impressionism &  .67   &  .58   & Neoclassicism &  .64   &  .60   \\
Orientalism &   .63  &  .63   & Pop-Art &  .62   &   .57  \\
P. Impressionism &  .61   &  .56   & Primitivism &  .61   &   .54  \\
Realism &  .60   &  .53   & Renaissance &  .64   &  .54   \\
Rococo &  .71   &   .59  & Romanticism &   .68  &  .59   \\
Surrealism &   .61  &  .50   & Symbolism &  .57   &  .59   \\
Ukiyo-e &  .66   &  .58   &  &     &     \\
\bottomrule
\end{tabular}%
\label{tab:rq2_masksga_sd}
\end{table}

\begin{table}[h]
\centering
\tiny 
\setlength{\tabcolsep}{2pt} 
\renewcommand{\arraystretch}{0.8} 
\caption{Full table of results on the Flux.1.Dev (Flux) case for the spatial and chromatic manipulations. The highlighted values are those contributing to the \textsc{PixelSGA} for each style. "G" stands for "grayscale", "B/W" refers to binary chromatic variation; "P. Impressionism" stands for "Post-Impressionism", "C. Realism" stands for "Contemporary Realism".}
\resizebox{\textwidth}{!}{%
\begin{tabular}{l l ccccccccc l l ccccccccc}
\toprule
\textbf{Style} & \textbf{Mode} & 224 & 112 & 56 & 28 & 14 & 7 & 4 & 2 & 1 &
\textbf{Style} & \textbf{Mode} & 224 & 112 & 56 & 28 & 14 & 7 & 4 & 2 & 1 \\
\midrule
\multirow{3}{*}{Academicism} & RGB & 1.00 & 1.00 & 1.00 & .99 & \textbf{.92} & \textbf{.79} & \textbf{.71} & \textbf{.64} & \textbf{.59} & \multirow{3}{*}{Art Nouveau} & RGB & 1.00 & 1.00 & 1.00 & .99 & \textbf{.87} & \textbf{.72} & \textbf{.64} & \textbf{.55} & .54 \\
& G & 1.00 & 1.00 & 1.00 & .96 & .82 & .73 & .63 & .54 & .53 & & G & 1.00 & 1.00 & .99 & .93 & .73 & .65 & .61 & .55 & \textbf{.56} \\
& B/W & .99 & .93 & .81 & .75 & .74 & .69 & .65 & .62 & .52 & & B/W & .98 & .95 & .78 & .70 & .61 & .55 & .58 & .49 & .49 \\
\midrule
\multirow{3}{*}{Baroque} & RGB & 1.00 & 1.00 & .99 & .99 & \textbf{.86} & \textbf{.73} & \textbf{.61} & \textbf{.59} & \textbf{.65} & \multirow{3}{*}{C. Realism} & RGB & 1.00 & 1.00 & 1.00 & .99 & \textbf{.96} & \textbf{.84} & \textbf{.74} & \textbf{.70} & \textbf{.64} \\
& G & 1.00 & 1.00 & .99 & .93 & .74 & .66 & .57 & .51 & .57 & & G & 1.00 & 1.00 & 1.00 & .97 & .89 & .81 & .69 & .61 & .59 \\
& B/W & .99 & .95 & .82 & .71 & .69 & .62 & .59 & .56 & .51 & & B/W & 1.00 & .99 & .93 & .84 & .81 & .66 & .65 & .52 & .58 \\
\midrule
\multirow{3}{*}{Cubism} & RGB & 1.00 & 1.00 & 1.00 & .99 & \textbf{.92} & \textbf{.70} & \textbf{.63} & \textbf{.63} & .61 & \multirow{3}{*}{Expressionism} & RGB & 1.00 & 1.00 & 1.00 & .98 & \textbf{.92} & \textbf{.71} & \textbf{.62} & \textbf{.61} & \textbf{.61} \\
& G & 1.00 & 1.00 & 1.00 & .95 & .78 & .64 & .56 & .58 & .60 & & G & 1.00 & 1.00 & 1.00 & .95 & .80 & .69 & .59 & .60 & .56 \\
& B/W & .99 & .97 & .87 & .75 & .66 & .58 & .58 & .54 & \textbf{.66} & & B/W & .99 & .96 & .90 & .78 & .63 & .64 & .54 & .55 & .58 \\
\midrule
\multirow{3}{*}{Impressionism} & RGB & 1.00 & 1.00 & 1.00 & .99 & \textbf{.95} & \textbf{.82} & \textbf{.73} & \textbf{.71} & \textbf{.60} & \multirow{3}{*}{Neoclassicism} & RGB & 1.00 & 1.00 & 1.00 & .99 & \textbf{.91} & \textbf{.75} & \textbf{.71} & \textbf{.61} & \textbf{.65} \\
& G & 1.00 & 1.00 & 1.00 & .97 & .86 & .77 & .72 & .65 & .52 & & G & 1.00 & 1.00 & 1.00 & .96 & .78 & .67 & .63 & .55 & .54 \\
& B/W & .99 & .98 & .94 & .86 & .78 & .65 & .63 & .54 & .57 & & B/W & .99 & .96 & .83 & .73 & .65 & .66 & .57 & .56 & .51 \\
\midrule
\multirow{3}{*}{Orientalism} & RGB & 1.00 & 1.00 & 1.00 & .99 & \textbf{.93} & \textbf{.76} & \textbf{.68} & .62 & \textbf{.67} & \multirow{3}{*}{Pop-Art} & RGB & 1.00 & 1.00 & 1.00 & 1.00 & \textbf{.94} & \textbf{.78} & \textbf{.69} & \textbf{.58} & \textbf{.62} \\
& G & 1.00 & 1.00 & 1.00 & .95 & .82 & .69 & .59 & \textbf{.64} & .61 & & G & 1.00 & 1.00 & 1.00 & .99 & .82 & .72 & .63 & .56 & .53 \\
& B/W & .96 & .91 & .81 & .68 & .64 & .61 & .60 & .52 & .36 & & B/W & 1.00 & 1.00 & .98 & .89 & .77 & .62 & .63 & .42 & .47 \\
\midrule
\multirow{3}{*}{P. Impressionism} & RGB & 1.00 & 1.00 & 1.00 & 1.00 & \textbf{.95} & \textbf{.79} & \textbf{.75} & \textbf{.68} & \textbf{.64} & \multirow{3}{*}{Primitivism} & RGB & 1.00 & 1.00 & 1.00 & .97 & \textbf{.87} & \textbf{.73} & \textbf{.66} & \textbf{.60} & \textbf{.64} \\
& G & 1.00 & 1.00 & 1.00 & .97 & .85 & .73 & .67 & .67 & .52 & & G & 1.00 & 1.00 & .99 & .95 & .80 & .66 & .53 & .56 & .56 \\
& B/W & .99 & .97 & .91 & .83 & .72 & .62 & .55 & .57 & .52 & & B/W & .99 & .93 & .83 & .70 & .58 & .58 & .59 & .53 & .51 \\
\midrule
\multirow{3}{*}{Realism} & RGB & 1.00 & 1.00 & 1.00 & .99 & \textbf{.93} & \textbf{.77} & \textbf{.65} & \textbf{.64} & \textbf{.62} & \multirow{3}{*}{Renaissance} & RGB & 1.00 & 1.00 & 1.00 & .99 & \textbf{.89} & \textbf{.75} & \textbf{.64} & .52 & \textbf{.63} \\
& G & 1.00 & 1.00 & 1.00 & .97 & .83 & .71 & .61 & .58 & .55 & & G & 1.00 & 1.00 & 1.00 & .96 & .79 & .71 & .60 & \textbf{.57} & .55 \\
& B/W & .98 & .95 & .87 & .79 & .70 & .63 & .62 & .60 & .50 & & B/W & .99 & .97 & .84 & .71 & .62 & .62 & .55 & .54 & .58 \\
\midrule
\multirow{3}{*}{Rococo} & RGB & 1.00 & 1.00 & 1.00 & .97 & \textbf{.87} & \textbf{.73} & \textbf{.64} & \textbf{.59} & \textbf{.65} & \multirow{3}{*}{Romanticism} & RGB & 1.00 & 1.00 & 1.00 & .99 & \textbf{.92} & \textbf{.82} & \textbf{.74} & \textbf{.64} & \textbf{.63} \\
& G & 1.00 & 1.00 & .98 & .94 & .78 & .72 & .62 & .54 & .52 & & G & 1.00 & 1.00 & 1.00 & .95 & .83 & .74 & .59 & .57 & .61 \\
& B/W & .97 & .97 & .84 & .69 & .66 & .60 & .59 & .57 & .52 & & B/W & .97 & .94 & .82 & .79 & .76 & .71 & .57 & .63 & .51 \\
\midrule
\multirow{3}{*}{Surrealism} & RGB & 1.00 & 1.00 & 1.00 & .96 & \textbf{.89} & \textbf{.72} & \textbf{.66} & \textbf{.61} & \textbf{.64} & \multirow{3}{*}{Symbolism} & RGB & 1.00 & 1.00 & 1.00 & .99 & \textbf{.87} & \textbf{.67} & \textbf{.62} & \textbf{.59} & \textbf{.62} \\
& G & 1.00 & 1.00 & .99 & .93 & .78 & .66 & .59 & .58 & .57 & & G & 1.00 & 1.00 & .99 & .93 & .76 & .61 & .53 & .58 & .51 \\
& B/W & .97 & .89 & .75 & .71 & .65 & .59 & .55 & .56 & .53 & & B/W & .94 & .87 & .75 & .65 & .63 & .63 & .49 & .45 & .51 \\
\midrule
\multirow{3}{*}{Ukiyo-e} & RGB & 1.00 & .99 & .99 & .96 & \textbf{.88} & \textbf{.77} & \textbf{.67} & .62 & \textbf{.72} & \multirow{3}{*}{} & &  &  &  &  &  &  &  &  &  \\
& G & .99 & 1.00 & .97 & .91 & .76 & .69 & .63 & .54 & .60 & & &  &  &  &  &  &  &  &  &  \\
& B/W & 1.00 & .98 & .93 & .82 & .71 & .63 & .63 & \textbf{.63} & .43 & & &  &  &  &  &  &  &  &  &  \\
\midrule
\end{tabular}%
}
\label{tab:rq2_pixelsga_flux}
\end{table}

\begin{table}[h]
\centering
\caption{Full table of results on the Flux.1.Dev (Flux) case with neutral prompts, considering the occlusion manipulations. All these values contribute to the \textsc{MaskSGA} for each style. "P. Impressionism" and "C. Realism" respectively stand for "Post-Impressionism" and "Contemporary Realism".}
\small
\setlength{\tabcolsep}{8pt} 
\small
\begin{tabular}{lcc|lcc}
\toprule
\textbf{Style} & \textbf{Mask} & \textbf{MaskNoBg} & \textbf{Style} & \textbf{Mask} & \textbf{MaskNoBg} \\
\midrule
Academicism &  .59   &  .53  & Art Nouveau &  .64   &  .52   \\
Baroque &  .59   &  .56   & C. Realism &  .69   &   .60  \\
Cubism &  .66   &  .54   & Expressionism &  .65   &  .58   \\
Impressionism &  .74   &  .61   & Neoclassicism &  .64   &  .57   \\
Orientalism &   .70  &  .60   & Pop-Art &  .75   &   .58  \\
P. Impressionism &  .71   &  .60   & Primitivism &  .69   &   .51  \\
Realism &  .64   &  .59   & Renaissance &  .60   &  .58   \\
Rococo &  .61   &   .58  & Romanticism &   .60  &  .58   \\
Surrealism &   .65  &  .55   & Symbolism &  .59   &  .56   \\
Ukiyo-e &  .75   &  .59   &  &     &     \\
\bottomrule
\end{tabular}%
\label{tab:rq2_masksga_flux}
\end{table}

\
\vspace{28cm}

For some general comparisons, we define SGA as the average of PixelSGA and MaskSGA as follows:

\begin{equation}
\textsc{SGA} = \frac{1}{2} \Big( \textsc{PixelSGA} + \textsc{MaskSGA} \Big).
\end{equation}

\begin{figure}[tb]
  \centering
  \includegraphics[width=.8\textwidth]{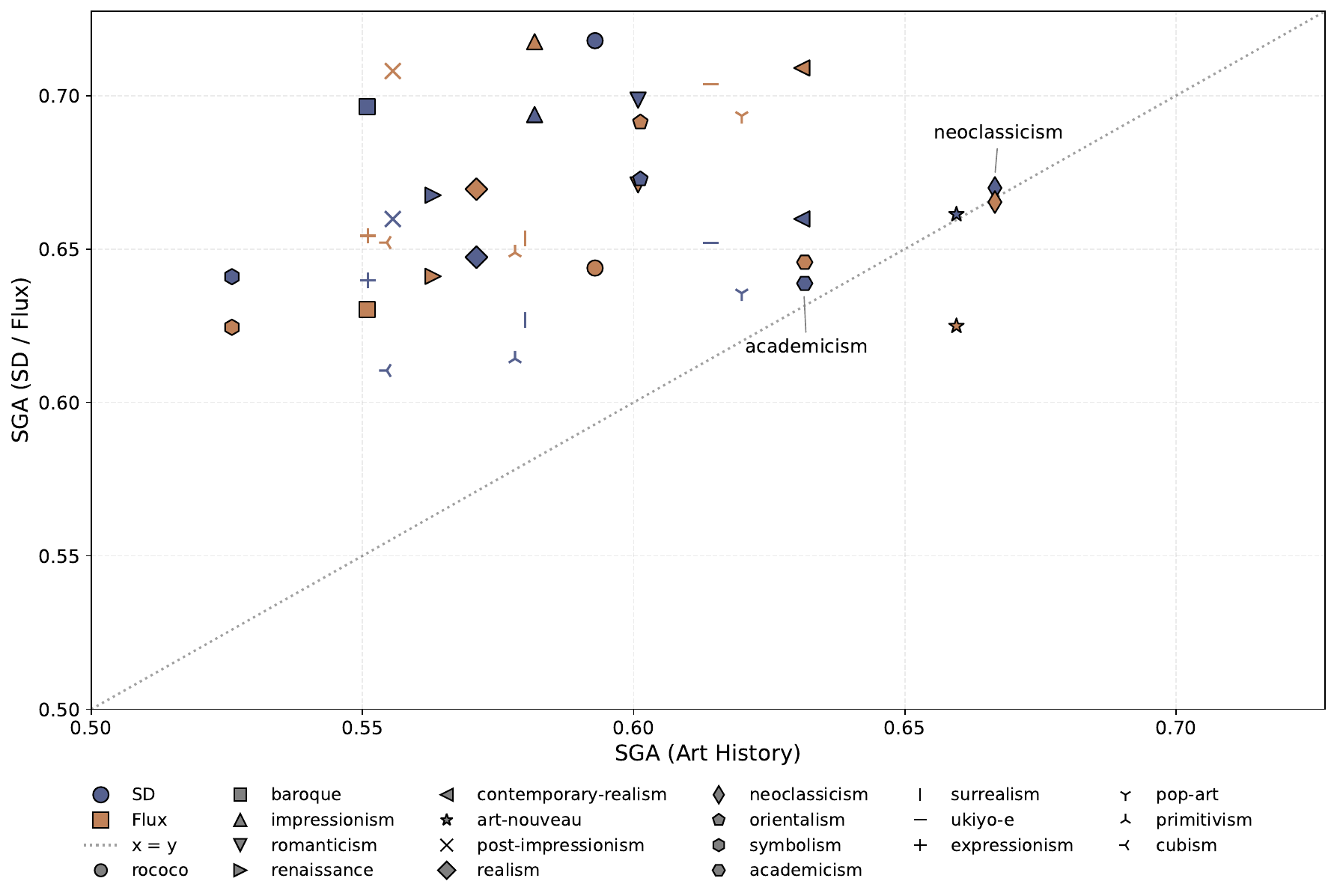}
  \caption{SGA scores per art style for historic paintings (x-axis) and images generated with Stable Diffusion 3.5 Medium (SD) and Flux.1-dev (Flux) (y-axis). Almost all dots are above the diagonal, indicating stronger gender artifacts for generated images than for historic paintings. Only for academicism and neoclassicism are the \textit{SD} and \textit{Flux} scores close together.
  }
  \label{fig:rq2}
\end{figure}

Figure \ref{fig:rq2} compares the SGA scores of art historical images with those of images generated by \textit{SD} and \textit{Flux}.
Interestingly, only Neoclassicism and Academicism show similar SGA values across the two models.

\begin{table}[h]
\centering
\small
\setlength{\tabcolsep}{4pt}
\renewcommand{\arraystretch}{0.9}
\caption{MaskSGA results from Art History to Text-to-Image Generation for both \textit{SD} and \textit{Flux}.}
\begin{tabular}{l l cc l l cc}
\toprule
\textbf{Style} & \textbf{Type} & BG & No BG & \textbf{Style} & \textbf{Type} & BG & No BG \\
\midrule
\multirow{2}{*}{Academicism} & \textit{SD} & \textbf{.73} & .58 & \multirow{2}{*}{Art-Nouveau} & \textit{SD} & .68 & .61 \\
 & \textit{Flux} & .71 & \textbf{.62} &  & \textit{Flux} & \textbf{.70} & \textbf{.63} \\
\midrule
\multirow{2}{*}{Baroque} & \textit{SD} & \textbf{.65} & \textbf{.63} & \multirow{2}{*}{Contemporary-Realism} & \textit{SD} & .55 & .59 \\
 & \textit{Flux} & .63 & .58 &  & \textit{Flux} & \textbf{.61} & \textbf{.61} \\
\midrule
\multirow{2}{*}{Cubism} & \textit{SD} & \textbf{.62} & .49 & \multirow{2}{*}{Expressionism} & \textit{SD} & .60 & .57 \\
 & \textit{Flux} & .54 & \textbf{.53} &  & \textit{Flux} & \textbf{.62} & \textbf{.60} \\
\midrule
\multirow{2}{*}{Impressionism} & \textit{SD} & .64 & .57 & \multirow{2}{*}{Neoclassicism} & \textit{SD} & \textbf{.73} & \textbf{.67} \\
 & \textit{Flux} & \textbf{.67} & \textbf{.67} &  & \textit{Flux} & .72 & .67 \\
\midrule
\multirow{2}{*}{Orientalism} & \textit{SD} & .53 & \textbf{.49} & \multirow{2}{*}{Pop-Art} & \textit{SD} & .56 & .49 \\
 & \textit{Flux} & \textbf{.59} & .45 &  & \textit{Flux} & \textbf{.61} & \textbf{.62} \\
\midrule
\multirow{2}{*}{Post-Impressionism} & \textit{SD} & .59 & .59 & \multirow{2}{*}{Primitivism} & \textit{SD} & \textbf{.65} & .49 \\
 & \textit{Flux} & \textbf{.64} & \textbf{.64} &  & \textit{Flux} & .55 & \textbf{.51} \\
\midrule
\multirow{2}{*}{Realism} & \textit{SD} & .67 & .59 & \multirow{2}{*}{Renaissance} & \textit{SD} & .62 & \textbf{.61} \\
 & \textit{Flux} & \textbf{.70} & \textbf{.64} &  & \textit{Flux} & \textbf{.63} & .59 \\
\midrule
\multirow{2}{*}{Rococo} & \textit{SD} & \textbf{.74} & \textbf{.67} & \multirow{2}{*}{Romanticism} & \textit{SD} & \textbf{.70} & .61 \\
 & \textit{Flux} & .70 & .66 &  & \textit{Flux} & .69 & \textbf{.61} \\
\midrule
\multirow{2}{*}{Surrealism} & \textit{SD} & .59 & .50 & \multirow{2}{*}{Symbolism} & \textit{SD} & .55 & \textbf{.56} \\
 & \textit{Flux} & \textbf{.61} & \textbf{.56} &  & \textit{Flux} & \textbf{.60} & .56 \\
\midrule
\multirow{2}{*}{Ukiyo-E} & \textit{SD} & .68 & \textbf{.60} &  & \textit{SD} &  &  \\
 & \textit{Flux} & \textbf{.72} & .60 &  & \textit{Flux} &  &  \\
\bottomrule
\end{tabular}%
\label{tab:rq3_masksga}
\end{table}

\subsection{Gender Artifacts from Art History to Text-to-Image Generation}
\label{sec:apx:rq3_results}

In Table \ref{tab:rq3_pixelsga}, we report additional results for the computation of \textsc{PixelSGA} for SD and Flux for this experiment. Even in this case, the values contributing to the metric (highlighted in bold) belong mostly to the RGB chromatic variation. In Table \ref{tab:rq3_masksga}, we report the results of the classifier for each style when masking the subject (Mask) and when masking also the background with a different color (MaskNoBg). All these values contribute to the computation of \textsc{MaskSGA} for each style.

\begin{table}[h]
\centering
\tiny
\setlength{\tabcolsep}{2pt}
\renewcommand{\arraystretch}{0.8}
\caption{Full table of results on the spatial and chromatic manipulations. The highlighted values are those contributing to the \textsc{PixelSGA} for each style. "G" stands for "grayscale", "B/W" refers to binary chromatic variation; "P. Impressionism" stands for "Post-Impressionism", "C. Realism" stands for "Contemporary Realism".}
\resizebox{\textwidth}{!}{%
\begin{tabular}{l l ccccccccc l l ccccccccc}
\toprule
\textbf{Style} & \textbf{Mode} & 224 & 112 & 56 & 28 & 14 & 7 & 4 & 2 & 1 & \textbf{Style} & \textbf{Mode} & 224 & 112 & 56 & 28 & 14 & 7 & 4 & 2 & 1 \\\\
\midrule
\multirow{3}{*}{Academicism} & RGB & .98 & \textbf{.98} & \textbf{.96} & \textbf{.94} & \textbf{.82} & \textbf{.77} & \textbf{.74} & \textbf{.68} & \textbf{.64} & \multirow{3}{*}{Academicism F} & RGB & \textbf{.99} & \textbf{.99} & \textbf{.96} & \textbf{.92} & \textbf{.86} & \textbf{.80} & \textbf{.74} & \textbf{.73} & \textbf{.65} \\
 & G & \textbf{.99} & .98 & .95 & .87 & .78 & .70 & .68 & .64 & .56 &  & G & .98 & .98 & .95 & .87 & .78 & .71 & .67 & .64 & .58 \\
 & B/W & .94 & .89 & .81 & .72 & .64 & .59 & .62 & .60 & .49 &  & B/W & .92 & .87 & .78 & .73 & .69 & .68 & .65 & .61 & .52 \\
\midrule
\multirow{3}{*}{Art-Nouveau} & RGB & \textbf{.98} & \textbf{.97} & \textbf{.96} & \textbf{.89} & \textbf{.77} & \textbf{.73} & \textbf{.69} & \textbf{.66} & \textbf{.64} & \multirow{3}{*}{Art-Nouveau F} & RGB & .96 & \textbf{.96} & \textbf{.94} & \textbf{.87} & \textbf{.80} & \textbf{.76} & \textbf{.73} & \textbf{.70} & \textbf{.66} \\
 & G & .98 & .97 & .95 & .87 & .74 & .68 & .66 & .60 & .57 &  & G & \textbf{.96} & .95 & .93 & .86 & .76 & .69 & .67 & .62 & .56 \\
 & B/W & .95 & .94 & .84 & .71 & .63 & .57 & .56 & .52 & .50 &  & B/W & .94 & .90 & .80 & .67 & .61 & .58 & .54 & .54 & .53 \\
\midrule
\multirow{3}{*}{Baroque} & RGB & \textbf{.97} & \textbf{.97} & \textbf{.96} & \textbf{.90} & \textbf{.81} & \textbf{.71} & \textbf{.67} & \textbf{.62} & \textbf{.56} & \multirow{3}{*}{Baroque F} & RGB & \textbf{.97} & \textbf{.96} & \textbf{.94} & \textbf{.91} & \textbf{.79} & \textbf{.67} & \textbf{.63} & .60 & \textbf{.62} \\
 & G & .97 & .96 & .95 & .86 & .71 & .65 & .62 & .60 & .55 &  & G & .96 & .96 & .93 & .86 & .71 & .65 & .61 & \textbf{.61} & .59 \\
 & B/W & .94 & .88 & .78 & .71 & .65 & .59 & .56 & .60 & .47 &  & B/W & .92 & .89 & .74 & .67 & .64 & .59 & .60 & .50 & .41 \\
\midrule
\multirow{3}{*}{Contemporary-Realism} & RGB & .88 & \textbf{.90} & \textbf{.86} & \textbf{.81} & \textbf{.71} & .64 & \textbf{.54} & .59 & .50 & \multirow{3}{*}{Contemporary-Realism F} & RGB & \textbf{.89} & \textbf{.88} & \textbf{.86} & \textbf{.81} & \textbf{.75} & \textbf{.63} & \textbf{.65} & \textbf{.58} & \textbf{.64} \\
 & G & \textbf{.90} & .86 & .85 & .78 & .64 & \textbf{.69} & .52 & \textbf{.64} & \textbf{.53} &  & G & .89 & .86 & .84 & .77 & .68 & .62 & .52 & .52 & .49 \\
 & B/W & .83 & .82 & .75 & .60 & .61 & .52 & .50 & .56 & .50 &  & B/W & .81 & .77 & .70 & .64 & .58 & .59 & .57 & .49 & .50 \\
\midrule
\multirow{3}{*}{Cubism} & RGB & \textbf{.85} & \textbf{.81} & \textbf{.78} & \textbf{.66} & \textbf{.63} & .55 & .47 & .47 & \textbf{.51} & \multirow{3}{*}{Cubism F} & RGB & .85 & \textbf{.85} & \textbf{.87} & \textbf{.80} & \textbf{.74} & \textbf{.60} & \textbf{.57} & \textbf{.58} & .48 \\
 & G & .83 & .80 & .75 & .64 & .59 & \textbf{.55} & \textbf{.56} & .50 & .45 &  & G & \textbf{.87} & \textbf{.85} & .83 & .79 & .60 & .57 & .54 & .45 & \textbf{.53} \\
 & B/W & .80 & .75 & .65 & .53 & .44 & .52 & .53 & \textbf{.51} & .50 &  & B/W & .82 & .81 & .68 & .57 & .50 & .55 & .51 & .53 & .50 \\
\midrule
\multirow{3}{*}{Expressionism} & RGB & .97 & \textbf{.96} & \textbf{.94} & \textbf{.86} & \textbf{.76} & \textbf{.64} & \textbf{.59} & \textbf{.56} & \textbf{.57} & \multirow{3}{*}{Expressionism F} & RGB & .97 & .96 & \textbf{.95} & \textbf{.90} & \textbf{.78} & \textbf{.66} & \textbf{.62} & \textbf{.58} & \textbf{.58} \\
 & G & \textbf{.98} & .96 & .92 & .82 & .68 & .59 & .56 & .55 & .49 &  & G & \textbf{.97} & \textbf{.97} & .93 & .85 & .72 & .64 & .60 & .54 & .47 \\
 & B/W & .95 & .91 & .79 & .66 & .59 & .53 & .52 & .50 & .53 &  & B/W & .93 & .89 & .80 & .68 & .62 & .59 & .56 & .51 & .52 \\
\midrule
\multirow{3}{*}{Impressionism} & RGB & \textbf{.94} & \textbf{.93} & \textbf{.92} & \textbf{.87} & \textbf{.78} & \textbf{.69} & \textbf{.65} & \textbf{.61} & \textbf{.55} & \multirow{3}{*}{Impressionism F} & RGB & \textbf{.98} & \textbf{.97} & \textbf{.96} & \textbf{.90} & \textbf{.80} & \textbf{.70} & \textbf{.69} & \textbf{.63} & \textbf{.59} \\
 & G & .93 & .92 & .89 & .80 & .70 & .63 & .63 & .59 & .50 &  & G & .98 & .96 & .94 & .83 & .72 & .69 & .64 & .60 & .57 \\
 & B/W & .86 & .83 & .71 & .63 & .59 & .55 & .50 & .54 & .48 &  & B/W & .92 & .88 & .78 & .65 & .60 & .60 & .57 & .52 & .50 \\
\midrule
\multirow{3}{*}{Neoclassicism} & RGB & \textbf{.99} & \textbf{.98} & \textbf{.97} & \textbf{.92} & \textbf{.86} & \textbf{.79} & \textbf{.75} & .69 & \textbf{.66} & \multirow{3}{*}{Neoclassicism F} & RGB & .97 & \textbf{.97} & \textbf{.96} & \textbf{.93} & \textbf{.88} & \textbf{.80} & \textbf{.77} & \textbf{.74} & \textbf{.65} \\
 & G & .99 & \textbf{.98} & .96 & .90 & .78 & .74 & .74 & \textbf{.70} & .58 &  & G & \textbf{.98} & .97 & .96 & .91 & .82 & .77 & .71 & .71 & .62 \\
 & B/W & .97 & .95 & .85 & .74 & .71 & .64 & .60 & .63 & .53 &  & B/W & .95 & .93 & .86 & .77 & .75 & .70 & .69 & .59 & .59 \\
\midrule
\multirow{3}{*}{Orientalism} & RGB & \textbf{.96} & .96 & .91 & \textbf{.81} & \textbf{.70} & \textbf{.62} & .57 & \textbf{.52} & \textbf{.53} & \multirow{3}{*}{Orientalism F} & RGB & .94 & \textbf{.96} & .90 & \textbf{.82} & \textbf{.69} & .59 & \textbf{.65} & \textbf{.62} & \textbf{.58} \\
 & G & .95 & \textbf{.96} & \textbf{.92} & .81 & .66 & .55 & \textbf{.58} & .50 & .49 &  & G & \textbf{.95} & .92 & \textbf{.91} & .77 & .65 & .58 & .62 & .56 & .51 \\
 & B/W & .88 & .84 & .65 & .61 & .60 & .49 & .52 & .44 & .50 &  & B/W & .86 & .76 & .68 & .63 & .51 & \textbf{.59} & .55 & .52 & .50 \\
\midrule
\multirow{3}{*}{Pop-Art} & RGB & .95 & \textbf{.96} & \textbf{.92} & \textbf{.84} & \textbf{.68} & .62 & \textbf{.53} & \textbf{.53} & \textbf{.53} & \multirow{3}{*}{Pop-Art F} & RGB & .90 & .87 & \textbf{.90} & \textbf{.87} & \textbf{.80} & \textbf{.65} & .55 & .51 & .52 \\
 & G & \textbf{.96} & .93 & .90 & .74 & .64 & \textbf{.63} & .50 & .51 & .52 &  & G & .88 & \textbf{.88} & .88 & .84 & .65 & .60 & .53 & \textbf{.61} & \textbf{.54} \\
 & B/W & .93 & .91 & .83 & .64 & .53 & .49 & .45 & .53 & .50 &  & B/W & \textbf{.91} & \textbf{.88} & .84 & .71 & .66 & .60 & \textbf{.58} & .55 & .50 \\
\midrule
\multirow{3}{*}{Post-Impressionism} & RGB & \textbf{.97} & \textbf{.97} & \textbf{.95} & \textbf{.88} & \textbf{.76} & \textbf{.65} & \textbf{.62} & .57 & \textbf{.58} & \multirow{3}{*}{Post-Impressionism F} & RGB & .97 & \textbf{.96} & \textbf{.95} & \textbf{.88} & \textbf{.76} & \textbf{.66} & \textbf{.62} & \textbf{.58} & \textbf{.62} \\
 & G & .97 & .97 & .92 & .81 & .69 & .60 & .58 & .54 & .56 &  & G & \textbf{.98} & .96 & .93 & .82 & .69 & .63 & .56 & .58 & .57 \\
 & B/W & .93 & .89 & .77 & .67 & .57 & .54 & .55 & \textbf{.58} & .53 &  & B/W & .93 & .89 & .76 & .66 & .57 & .51 & .52 & .51 & .52 \\
\midrule
\multirow{3}{*}{Primitivism} & RGB & .91 & \textbf{.93} & \textbf{.90} & \textbf{.86} & \textbf{.74} & \textbf{.60} & \textbf{.60} & \textbf{.60} & .58 & \multirow{3}{*}{Primitivism F} & RGB & \textbf{.94} & \textbf{.90} & \textbf{.92} & \textbf{.85} & \textbf{.77} & .63 & .60 & .56 & .54 \\
 & G & \textbf{.92} & .91 & .89 & .78 & .64 & .55 & .53 & .58 & \textbf{.63} &  & G & .92 & .88 & .87 & .79 & .66 & \textbf{.64} & \textbf{.62} & \textbf{.60} & \textbf{.62} \\
 & B/W & .85 & .84 & .69 & .55 & .61 & .51 & .50 & .52 & .50 &  & B/W & .87 & .80 & .72 & .54 & .57 & .51 & .46 & .57 & .50 \\
\midrule
\multirow{3}{*}{Realism} & RGB & \textbf{.98} & \textbf{.97} & \textbf{.96} & \textbf{.91} & \textbf{.81} & \textbf{.69} & \textbf{.66} & \textbf{.62} & .60 & \multirow{3}{*}{Realism F} & RGB & .97 & \textbf{.96} & \textbf{.96} & \textbf{.91} & \textbf{.83} & \textbf{.72} & \textbf{.67} & \textbf{.63} & \textbf{.61} \\
 & G & .98 & .97 & .95 & .88 & .73 & .65 & .62 & .58 & .55 &  & G & \textbf{.97} & .96 & .95 & .87 & .74 & .65 & .61 & .59 & .53 \\
 & B/W & .95 & .93 & .83 & .71 & .62 & .57 & .56 & .51 & \textbf{.67} &  & B/W & .92 & .89 & .79 & .69 & .64 & .59 & .57 & .54 & .49 \\
\midrule
\multirow{3}{*}{Renaissance} & RGB & .99 & \textbf{.99} & .96 & \textbf{.92} & \textbf{.82} & \textbf{.71} & \textbf{.64} & \textbf{.62} & \textbf{.60} & \multirow{3}{*}{Renaissance F} & RGB & \textbf{.98} & .97 & \textbf{.95} & \textbf{.90} & \textbf{.80} & \textbf{.70} & \textbf{.62} & \textbf{.59} & \textbf{.66} \\
 & G & \textbf{.99} & .98 & \textbf{.97} & .86 & .73 & .68 & .64 & .61 & .54 &  & G & .98 & \textbf{.97} & .94 & .86 & .73 & .65 & .62 & .56 & .60 \\
 & B/W & .97 & .93 & .81 & .72 & .67 & .61 & .57 & .58 & .49 &  & B/W & .94 & .91 & .79 & .68 & .68 & .60 & .55 & .52 & .49 \\
\midrule
\multirow{3}{*}{Rococo} & RGB & .98 & .98 & \textbf{.98} & \textbf{.96} & \textbf{.91} & \textbf{.79} & \textbf{.72} & \textbf{.71} & \textbf{.66} & \multirow{3}{*}{Rococo F} & RGB & .98 & \textbf{.98} & \textbf{.98} & \textbf{.97} & \textbf{.93} & \textbf{.80} & \textbf{.75} & \textbf{.71} & \textbf{.68} \\
 & G & \textbf{.98} & \textbf{.98} & .98 & .95 & .82 & .77 & .72 & .70 & .56 &  & G & \textbf{.98} & .98 & .97 & .95 & .85 & .76 & .73 & .68 & .55 \\
 & B/W & .98 & .96 & .90 & .79 & .75 & .68 & .67 & .67 & .55 &  & B/W & .97 & .96 & .88 & .81 & .79 & .73 & .67 & .59 & .50 \\
\midrule
\multirow{3}{*}{Romanticism} & RGB & .99 & .98 & \textbf{.97} & \textbf{.93} & \textbf{.85} & \textbf{.74} & \textbf{.68} & \textbf{.66} & \textbf{.60} & \multirow{3}{*}{Romanticism F} & RGB & .98 & .98 & \textbf{.97} & \textbf{.93} & \textbf{.85} & \textbf{.74} & \textbf{.72} & \textbf{.66} & \textbf{.61} \\
 & G & \textbf{.99} & \textbf{.98} & .96 & .89 & .77 & .73 & .68 & .65 & .55 &  & G & \textbf{.98} & \textbf{.98} & .96 & .88 & .78 & .70 & .67 & .62 & .57 \\
 & B/W & .95 & .92 & .83 & .73 & .68 & .59 & .61 & .57 & .51 &  & B/W & .94 & .90 & .79 & .73 & .67 & .64 & .64 & .56 & .49 \\
\midrule
\multirow{3}{*}{Surrealism} & RGB & .91 & .89 & \textbf{.86} & \textbf{.78} & \textbf{.66} & .56 & \textbf{.57} & .52 & \textbf{.52} & \multirow{3}{*}{Surrealism F} & RGB & .90 & .89 & .84 & \textbf{.80} & \textbf{.69} & .59 & .55 & \textbf{.56} & .55 \\
 & G & \textbf{.92} & \textbf{.90} & .85 & .72 & .61 & \textbf{.57} & .52 & \textbf{.56} & .46 &  & G & \textbf{.90} & \textbf{.90} & \textbf{.85} & .75 & .66 & \textbf{.60} & \textbf{.59} & .54 & \textbf{.58} \\
 & B/W & .85 & .82 & .72 & .58 & .50 & .50 & .50 & .51 & .50 &  & B/W & .85 & .79 & .71 & .61 & .52 & .59 & .55 & .52 & .50 \\
\midrule
\multirow{3}{*}{Symbolism} & RGB & \textbf{.95} & \textbf{.93} & .90 & \textbf{.80} & \textbf{.70} & \textbf{.63} & .59 & \textbf{.59} & \textbf{.59} & \multirow{3}{*}{Symbolism F} & RGB & \textbf{.94} & .91 & \textbf{.91} & \textbf{.79} & \textbf{.68} & \textbf{.59} & \textbf{.60} & \textbf{.57} & .55 \\
 & G & .95 & \textbf{.93} & \textbf{.91} & .71 & .64 & .60 & \textbf{.59} & .56 & .53 &  & G & .93 & \textbf{.93} & .88 & .76 & .61 & .57 & .58 & .48 & \textbf{.56} \\
 & B/W & .91 & .85 & .72 & .60 & .53 & .53 & .54 & .50 & .45 &  & B/W & .85 & .77 & .63 & .56 & .52 & .53 & .55 & .45 & .42 \\
\midrule
\multirow{3}{*}{Ukiyo-E} & RGB & .94 & \textbf{.91} & \textbf{.86} & \textbf{.74} & \textbf{.66} & \textbf{.64} & .54 & .53 & .53 & \multirow{3}{*}{Ukiyo-E F} & RGB & .93 & \textbf{.93} & .89 & \textbf{.87} & .75 & \textbf{.71} & .62 & .54 & \textbf{.59} \\
 & G & \textbf{.95} & .91 & .81 & .69 & .61 & .55 & \textbf{.55} & .50 & .50 &  & G & .93 & .92 & \textbf{.89} & .78 & \textbf{.78} & .64 & .65 & \textbf{.60} & .51 \\
 & B/W & .88 & .85 & .68 & .64 & .62 & .55 & .54 & \textbf{.56} & \textbf{.55} &  & B/W & \textbf{.94} & .91 & .81 & .68 & .67 & .63 & \textbf{.67} & .57 & .52 \\
\bottomrule
\end{tabular}%
}
\label{tab:rq3_pixelsga}
\end{table}

\subsection{Robustness Analysis}
\label{sec:apx:robustness}

\begin{figure}[tb]
    \centering
    \includegraphics[width=\linewidth]{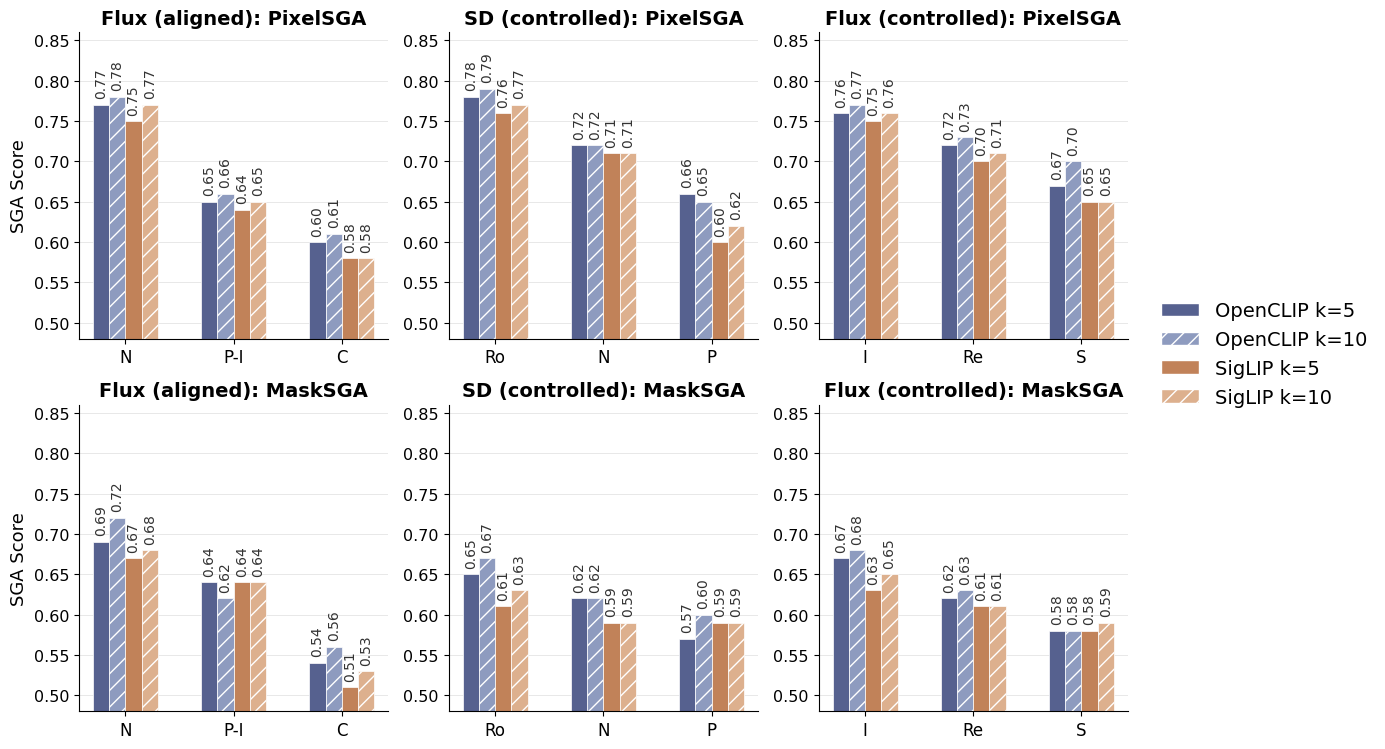}
    \caption{Robustness analysis of our metrics (PixelSGA and MaskSGA) to the choice of embedding model. Here we report results on images generated with Stable Diffusion (SD) and Flux, for both semantically aligned and controlled prompts. Style abbreviations: N (Neoclassicism), P-I (Post-Impressionism), C (Cubism), R (Rococo), P (Primitivism), I (Impressionism), R (Realism), S (Symbolism). See also Figure~\ref{fig:robustness_main} in the main text.}
    \label{fig:robustness_apx}
\end{figure}

As $k=5$ and CLIP are particular choices that should not influence the results qualitatively, we perform additional experiments on some of the styles of both historical and generated images. As reported in Figure \ref{fig:robustness_apx}, the results of different metrics are indeed robust to changes of $k$ or the chosen embedding model. For each setting, we select three styles that score high, medium, and low on both metrics. We consider Neoclassicism, Post-Impressionism and Cubism for \textit{Flux} with semantically aligned prompts; Realism, Neoclassicism and Primitivism for \textit{SD} with structured prompts; and finally Impressionism, Realism and Symbolism for \textit{Flux} with structured prompts. In all cases, the orderings between styles are preserved.

\subsection{Computational Resources}
\label{sec:apx:computation}

We estimate the total compute to be approximately 300 GPU-hours on 
Google Colab and 50 GPU-hours on the institutional cluster, accumulated 
over five months. Individual runs typically required 1-2 hours on Colab 
and 4-8 hours on the institutional cluster, depending on the task. 
Additionally, a few preliminary experiments conducted during the initial 
implementation and debugging phase did not contribute to the final results 
reported in the paper.


\clearpage
\section*{NeurIPS Paper Checklist}

\begin{enumerate}

\item {\bf Claims}
    \item[] Question: Do the main claims made in the abstract and introduction accurately reflect the paper's contributions and scope?
    \item[] Answer: \answerYes{} 
    \item[] Justification: In Section \ref{sec:dataset}, we introduce the \textsc{StyleGender} dataset, providing details regardings its three different sources. We further introduce the Set Gender Artifacts (SGA) metric in Section \ref{sec:metric}, and we report the results of our analysis in Section \ref{sec:experiments}.

\item {\bf Limitations}
    \item[] Question: Does the paper discuss the limitations of the work performed by the authors?
    \item[] Answer: \answerYes{} 
    \item[] Justification: The limitations are discussioned in Section \ref{sec:discussion}.

\item {\bf Theory assumptions and proofs}
    \item[] Question: For each theoretical result, does the paper provide the full set of assumptions and a complete (and correct) proof?
    \item[] Answer: \answerNA{} 
    \item[] Justification: The paper does not include theoretical results.

    \item {\bf Experimental result reproducibility}
    \item[] Question: Does the paper fully disclose all the information needed to reproduce the main experimental results of the paper to the extent that it affects the main claims and/or conclusions of the paper (regardless of whether the code and data are provided or not)?
    \item[] Answer: \answerYes{} 
    \item[] Justification: The details to reproduce our results are provided in Section \ref{sec:metric}, as well as in Section \ref{sec:experiments}. Moreover, further implementation details (\textit{e.g.}, the parameters of the bounding boxes for the computation of \textsc{MaskSGA}) are also provided throughout the Appendix (\textit{e.g.}, in Section \ref{sec:apx:bb}).

\item {\bf Open access to data and code}
    \item[] Question: Does the paper provide open access to the data and code, with sufficient instructions to faithfully reproduce the main experimental results, as described in supplemental material?
    \item[] Answer: \answerYes{} 
    \item[] Justification: Both data and code are submitted. We have done the best possible effort to ensure reproducibility of the results.

\item {\bf Experimental setting/details}
    \item[] Question: Does the paper specify all the training and test details (e.g., data splits, hyperparameters, how they were chosen, type of optimizer) necessary to understand the results?
    \item[] Answer: \answerYes{} 
    \item[] Justification: At the beginning of Section \ref{sec:experiments}, we discuss the choice of the embedding model, and settings for the kNN classifier.

\item {\bf Experiment statistical significance}
    \item[] Question: Does the paper report error bars suitably and correctly defined or other appropriate information about the statistical significance of the experiments?
    \item[] Answer: \answerYes{} 
    \item[] Justification: Statistical significance of our results is reported in the monotonicity sanity check (Section \ref{sec:experiments} and Section \ref{sec:apx:mono}), as well in the robustness check reported for each experiment and in the discussion of the last experiment in Section \ref{sec:apx:robustness}. 

\item {\bf Experiments compute resources}
    \item[] Question: For each experiment, does the paper provide sufficient information on the computer resources (type of compute workers, memory, time of execution) needed to reproduce the experiments?
    \item[] Answer: \answerYes{} 
    \item[] Justification: Details are provided in Section \ref{sec:discussion} and in Appendix \ref{sec:apx:computation}. 
    
\item {\bf Code of ethics}
    \item[] Question: Does the research conducted in the paper conform, in every respect, with the NeurIPS Code of Ethics \url{https://neurips.cc/public/EthicsGuidelines}?
    \item[] Answer: \answerYes{} 
    \item[] Justification: No human subjects were included in this research. The art historical images are downloaded from WikiArt, respecting the copyright of such images (most of them are in the public domain). The generated images do not represent any existing person.

\item {\bf Broader impacts}
    \item[] Question: Does the paper discuss both potential positive societal impacts and negative societal impacts of the work performed?
    \item[] Answer: \answerYes{} 
    \item[] Justification: Positive impact of this work is discussed throughout the paper. In Section \ref{sec:discussion}, we also report potential negative implications.
    
\item {\bf Safeguards}
    \item[] Question: Does the paper describe safeguards that have been put in place for responsible release of data or models that have a high risk for misuse (e.g., pre-trained language models, image generators, or scraped datasets)?
    \item[] Answer: \answerYes{} 
    \item[] Justification: In Section \ref{sec:discussion}, we elaborate on possible gender stereotypes and we inform the choice of sharing the dataset under a CC BY-NC-SA that restrict commercial usage.

\item {\bf Licenses for existing assets}
    \item[] Question: Are the creators or original owners of assets (e.g., code, data, models), used in the paper, properly credited and are the license and terms of use explicitly mentioned and properly respected?
    \item[] Answer: \answerYes{} 
    \item[] Justification: As specified in Section \ref{sec:dataset} and Appendix \ref{sec:apx:wiki} for the art historical images.

\item {\bf New assets}
    \item[] Question: Are new assets introduced in the paper well documented and is the documentation provided alongside the assets?
    \item[] Answer: \answerYes{} 
    \item[] Justification: We describe our dataset \textsc{StyleGender} in Section \ref{sec:dataset} and provide additional details in Appendix \ref{sec:apx:dataset}.

\item {\bf Crowdsourcing and research with human subjects}
    \item[] Question: For crowdsourcing experiments and research with human subjects, does the paper include the full text of instructions given to participants and screenshots, if applicable, as well as details about compensation (if any)? 
    \item[] Answer: \answerNA{} 
    \item[] Justification: Non-applicable.

\item {\bf Institutional review board (IRB) approvals or equivalent for research with human subjects}
    \item[] Question: Does the paper describe potential risks incurred by study participants, whether such risks were disclosed to the subjects, and whether Institutional Review Board (IRB) approvals (or an equivalent approval/review based on the requirements of your country or institution) were obtained?
    \item[] Answer: \answerNA{} 
    \item[] Justification: Non-applicable.

\item {\bf Declaration of LLM usage}
    \item[] Question: Does the paper describe the usage of LLMs if it is an important, original, or non-standard component of the core methods in this research? Note that if the LLM is used only for writing, editing, or formatting purposes and does \emph{not} impact the core methodology, scientific rigor, or originality of the research, declaration is not required.
    \item[] Answer: \answerYes{} 
    \item[] Justification: In Section \ref{sec:dataset}, we describe how LLMs have allowed the translation of artistic images from art history to T2I generation. All the used LLMs are properly communicated. In addition, in Section \ref{sec:apx:qwen}, we also share how we have manually corrected captioning errors.

\end{enumerate}

\end{document}